\documentclass[]{article}
\usepackage[affil-it]{authblk}
\usepackage[a4paper, total={6.5in, 8in}]{geometry}
\usepackage{graphicx}
\usepackage{pgfplots}
\usepackage{booktabs} 
\usepackage{graphicx}
\usepackage{pgfplots}
\usepackage[all]{nowidow}
\usepackage[utf8]{inputenc}
\usepackage{tikz}
\usepackage{fancyhdr}
\usepackage{dblfloatfix}
\usepackage{float}
\usepackage{textgreek}
\usepackage{multicol}
\usepackage{algorithm}
\usepackage{relsize}
\usepackage{mathtools}

\usepackage{hyperref}
\usepackage{subfig}
\usepackage{verbatim}
\usepackage{algorithmic}
\usepackage[inline]{enumitem} 
\usepackage{setspace}

	\title{\textbf{Cluster Weighted Model Based on TSNE algorithm for High-Dimensional Data}}

\author{Kehinde Olobatuyi%
	\thanks{Electronic address: \texttt{k.olobatuyi@campus.unimib.it}; Corresponding author}}
\affil{Statistics and Mathematical Finance,\\ University of Milano-Bicocca, Italy}

\date{}

\definecolor{blue}{HTML}{1F77B4}
\definecolor{orange}{HTML}{FF7F0E}
\definecolor{green}{HTML}{2CA02C}

\pgfplotsset{compat=1.14}

\newcommand\MEETtitle[1]{\Large \bf \hskip2.25pc \parbox{.8\textwidth}{ \noindent%
		\large \bf \begin{center} #1 \end{center}\rm } \vskip.1in \rm\normalsize }

\newcommand\MEETauthor[1]{\hskip2.25pc \parbox{.8\textwidth}{ \noindent%
		\normalsize \bf \begin{center} #1 \end{center}\rm } \vskip-1pc }

\let\title\MEETtitle
\let\author\MEETauthor
\let\affil\MEETaddress

\begin{document}
	
	\maketitle
	
	
\begin{abstract}
Similar to many Machine Learning models, both accuracy and speed of the Cluster weighted models (CWMs) can be hampered by high-dimensional data, leading to previous works on a parsimonious technique to reduce the effect of "Curse of dimensionality" on mixture models. In this work, we review the background study of the cluster weighted models (CWMs). We further show that parsimonious technique is not sufficient for mixture models to thrive in the presence of huge high-dimensional data. We discuss a heuristic for detecting the hidden components by choosing the initial values of location parameters using the default values in the "FlexCWM" R package. We introduce a dimensionality reduction technique called T-distributed stochastic neighbor embedding (TSNE) to enhance the parsimonious CWMs in high-dimensional space. Originally, CWMs are suited for regression but for classification purposes, all multi-class variables are transformed logarithmically with some noise. The parameters of the model are obtained via expectation maximization algorithm. The effectiveness of the discussed technique is demonstrated using real data sets from different fields. 
		
		\medskip
		
		
	\end{abstract}	
	

	\section{Background History}
	Efficient dimension reduction is required to uncover the hidden patterns of information in real data such as engineering data. Dimension reduction can be used to convert data sets containing millions of functions into a practicable spaces for effective processing and analysis. Unsupervised learning is the main technique for dimensionality reduction. Conventional dimensionality reduction approaches can be integrated with statistical analysis to improve the performance of the high-dimensional data [\cite{rehmanetal2016}]. \par Many dimensionality reduction techniques such as Principal component analysis (PCA) and T-distributed Stochastic Neighbor Embedding (TSNE) have been developed by statistical and artificial intelligence communities. In the recent years, there has been a rising interest in PCA mixture models. Mixture models provide an important framework for modeling complex data with a weighted component distribution. Due to their high flexibility and efficiency, they are used widely in many fields such as machine learning, image processing, and data mining. However, implementation in high-dimensional spaces are restricted by practical considerations because the component distributions are formalized as probability density functions. PCA mixture models are mixture-of-experts technique that model a nonlinear distribution through a combination of local linear sub-models with a fair sample distribution [\cite{jinetal2004}]. For more information on PCA mixture models, interested readers should look up the following [\cite{kimetal2003},\cite{xuetal2014},\cite{kutluketal2016}]. However, the main limitation is the linearity of PCA. PCA creates linear combinations of the existing features, so it fails to capture the nonlinearity in the features, thereby it is not able to interpret complex polynomial relationship between features. Therefore, PCA generally performs poorly if the relationship between the variables is nonlinear which may be inevitably present in high-dimensional data.
	\section{Methods}
	\subsection{CWMs-TSNE for High-dimensional data}\label{sec:3.4}
	Finite mixture models have been widely and successfully applied in many fields such as biological, genetics, medicine, psychiatry, economics, engineering, marketing, astronomy, among many other fields in the biological, physical, and social sciences. In these applications, finite mixture models underpin a variety of techniques in major areas of statistics, including latent class analysis, discriminant analysis, image analysis, and survival analysis, in addition to their more direct role in data analysis and inference of providing descriptive models for distributions. \par However, most of the finite mixture models assume the \textit{assignment independence} which implies that the probability for a point to be generated by one of the cluster must be the same for all the covariate values \textbf{x}. On the other hand the assignment of data point into the cluster must be independent of the covariates [\cite{Hennig2000}]. The cluster membership is determined by the covariate values. There are two reasonable models for linear regression clusters that do not assume assignment independence. One strategy used is to replace the fixed covariates by covariate distributions that are allowed to differ between the clusters. This is also similar to the evolution of cluster weighted models. It assumes the varying covariates with a parameterized family of distributions. This solves the problem of \textit{assignment independence} i.e. the covariate distributions of the mixture components is unique across the cluster. In the framework of mixture models with varying covariates, the cluster weighted model [CWM; \cite{Gershenfeld1997}], is given by the equation
	\begin{equation}
		p(y,\vec{x}) = \mathlarger\sum_{g=1}^{G} \pi_g p(y,\vec{x}|\mathcal{D}_g) = \mathlarger \sum_{g=1}^{G} \pi_g p(y|\vec{x},\mathcal{D}_g)p(\vec{x}|\mathcal{D}_g),
		\label{3.7}
	\end{equation}
	also called the saturated mixture regression model [\cite{Wedel2002}], constitutes a reference approach to model the joint density. In Equation (\ref{3.7}), normality of both $p(y|\vec{x},\mathcal{D}_g)$ and $p(\vec{x}|\mathcal{D}_g)$ is commonly assumed [\cite{Punzo2012}; \cite{Gershenfeld1997}].
	In this paper, we will focus mainly on the application of CWMs to high-dimensional data. The data considered in this paper ranges from the tens to hundreds of features. Like any machine learning classifiers, CWMs clustering performance can be hindered by the redundancies in the feature space of the data. Moreover, the computation speed reduces exponentially with increase in dimensionality. We hereby present CWMs in the presence of high-dimensional data. We begin the discussion with an interplay between t-distributed stochastic neighbor embedding and CWMs for clustering high-dimensional data.
	\subsection{T-distributed Stochastic Neighbor Embedding Technique}
	The stochastic neighbor embedding (SNE) was first introduced by \cite{hintonandroweis2002}. SNE aims to place the objects in a low-dimensional space in order to retain neighboring identity, and can be naturally extend to allow multiple different low-dimensional images of each object, [\cite{hintonandroweis2002}]. As a dimensionality reduction technique, SNE can construct a reasonably good performance of visualizations, however, it is hindered by a complex cost function that is difficult to optimize. \cite{maatenandhinton2008} introduced a variation of SNE called TSNE. The aim of TSNE is to transform the high-dimensional data set $\mathrm{X} = (x_1,...,x_n)$ into low-dimensional data set $\mathrm{Y} = (y_1,...,y_n)$. tSNE is much easier to easier to optimize, and provides significantly better visualization by reducing the tendency to crowd points together in the center of the map, [\cite{maatenandhinton2008}]. The cost function employed in TSNE is different from that of SNE. TSNE employed a symmetric version of SNE as an alternative to mitigate the problem of the presence of outliers. The asymmetric SNE used in SNE is given as follows;
	\begin{equation}
		q_{ij} = \frac{\text{exp}(-||y_i - y_j||^2)}{\sum_{k \not= i}\text{exp}(-||y_k - y_j||^2)}
		\label{equ:3.270}
	\end{equation}
	where $q_{ij}$ is the pairwise similarities in the low dimensional map and the way to define the pairwise similarities in the high-dimensional space $p_{ij}$ is given by
	\begin{equation}
		p_{ij} = \frac{\text{exp}(-||x_i - x_j||^2)/2\sigma^2}{\sum_{k \not= i}\text{exp}(-||x_k - x_j||^2/2\sigma^2)}
		\label{equ:3.280}
	\end{equation}
	These equations are referred to as symmetric because it has $p_{ij} = p_{ji}$ and $q_{ij} = q_{ji}$ for $\forall i,j$. Another uniqueness with TSNE is that TSNE applies a Student t-distribution with degree of freedom $v = 1$, similar to Cauchy distribution as the heavy-tailed distribution in the low-dimensional space. The joint probabilities for the low-dimensional map $q_{ij}$ instead becomes 
	\begin{equation}
		q_{ij} = \frac{\big(1+||y_i - y_j||^2\big)^{-1}}{\sum_{k\not=i}\big(1+||y_i - y_j||^2\big)^{-1}}
		\label{equ:3.290}
	\end{equation}   
	The advantages of employing a Student t-distribution can be found in the \cite{maatenandhinton2008}. The ultimate goal of TSNE is to represent $p_{ij}$ by $q_{ij}$ as accurate as possible, so the cost function is given by
	\begin{equation}
		C = \text{KL}(P||Q) = \mathlarger \sum_i \mathlarger \sum_j p_{ij} \hspace*{0.02in} \text{log} \hspace{0.02in}\frac{p_{ij}}{q_{ij}}
		\label{equ:3.300}
	\end{equation}
	Gradient method is introduced for minimizing the cost function and the gradient has the form given by 
	\begin{equation}
		\frac{\partial C}{\partial y_i} = 4 \mathlarger \sum_j (p_{ij}-q_{ij})(y_i-y_j)\big(1+||y_i-y_j||^2\big)^{-1}
		\label{equ:3.311}
	\end{equation}
	Equation (\ref{equ:3.311}) can be interpreted as the summation of a resultant force pulling $y_i$ in the direction of $y_j$ or pushing it away depending on whether $j$ is observed as a neighbor of $i$. The gradient descent is initialized by sampling the map point $\mathrm{Y}^{(0)} = (y_1,...y_n)$ randomly from $\mathcal{N}(0, 10^{-4}I)$. A momentum is added to the gradient descent to speed up the optimization and avoid being stuck in local optimal. Finally, the gradient update is given by 
	\begin{equation}
		\mathrm{Y}^{(t)} = \mathrm{Y}^{(t-1)} + \zeta \frac{\partial C}{\partial \mathrm{Y}} + \alpha(t)(\mathrm{Y}^{(t-1)}-\mathrm{Y}^{(t-2)})
		\label{equ:3.321}
	\end{equation}
	where $\mathrm{Y}^{(t)}$ is the solution at the iteration $t$, $\zeta$ is the learning rate, and the $\alpha(t)$ is the momentum at iteration $t$.
	\subsection{General Formulation of CWMs}
	Let $(\mathbf{X}, Y)$ be a pair of random vector $\mathbf{X}$ and random variable $Y$ defined on $\mathcal{D}$ with joint probability $p(\mathbf{x}, y)$, where $\mathbf{X}$ is a $d$-dimensional input vector with values in some space $\mathcal{X} \subseteq \mathrm{R}^d$ and $Y$ is a response variable having values in $\mathcal{Y} \subseteq \mathrm{R}$. The set of all model parameters is denoted $\mathbf{\Theta} = (\mathbf{\omega}, \mathbf{\mu}, \mathbf{\Sigma}, \pi)$.
	To begin with, we state that $\omega \in \mathrm{R}^{d \times G}$ denotes the weight of the local model, location parameter $\vec{\mu} \in \mathrm{R}^{d \times G}$, $G$ is the number of groups, $\vec{\Sigma}$ is the positive definite covariance matrix, and the $\pi$ is the mixing distribution with some constraints such as $\sum_g \pi_g = 1$ and $\pi_g > 0.$
	\vspace*{0.05in}
	\bigskip
	\par \noindent Generally, CWMs are written as a sum 
	\begin{equation}
		p(\mathbf{x}, y) = \sum_{g=1}^{G} p_g(\mathbf{x}, y),
		\label{equ:3.9}
	\end{equation}  
	where $g$ enumerates the clusters, and $p_g(y, \mathbf{x})$ is a density of a specific form discussed below. The total number of clusters $G$ must be chosen beforehand and can be selected based on the information criteria. The density $p_g(\mathbf{x}, y)$ is written as
	\begin{equation}
		\textit{p}(\mathbf{y},\mathbf{x})\hspace{.2cm} = \sum^G_{g = 1}\hspace{0.1cm}\textit{p}(\textbf{y},\textbf{x}, z_g)\hspace{0.1cm}
		\label{equ:3.10}
	\end{equation}
	The density $p_g(\mathbf{x},y)$ is written as
	\begin{equation}
		p_g(\mathbf{x},y) = \hspace{0.1cm}\textit{p}(\textbf{y}|\textbf{x}, z_g)\hspace{0.1cm}\textit{p}(\textbf{x}|z_g)\hspace{0.1cm}\pi_g
		\label{equ:3.11}
	\end{equation}
	Where $\textit{p}(z_g) = \pi_g$.
	The terms in equation (\ref{equ:3.11}) have the following interpretation:
	\vskip 0.03in
	~\\
	{\bf Cluster Weights:} The cluster weight $\pi_g \in [0,1]$ denotes the amount of data described by the cluster $g$. The $\pi_g$ are chosen subject to the constraint
	\begin{equation}
		\sum_{g=1}^{G}\pi_g = 1
		\label{equ:3.12}
	\end{equation} 
	{\bf Probability of inputs:} The density $p_m(\mathbf{x}) = p(\mathbf{x}|z_g)$ describes the domain of influence of cluster $g$, that is, the distribution of inputs $\mathbf{x}$ around the cluster. They are chosen with assumption as Gaussian densities, i.e.
	\begin{equation}
		\textit{p}(\mathbf{x}|z_g)\sim\mathcal{N}(\vec{\mu}_g, \vec{\Sigma}_g)
		\label{equ:3.13}
	\end{equation}
	\begin{equation}
		p_g(\mathbf{x}) = \frac{|\vec{\Sigma}_g^{-1}|^{1/2}}{(2\pi)^{d/2}} \exp(-\frac{1}{2}(\mathbf{x} - \vec{\mu}_g)^T\vec{\Sigma}_g^{-1}(\vec{x}-\vec{\mu}_g)
		\label{equ:3.14}
	\end{equation}
	with mean $\vec{\mu}_g$ and covariance matrix $\vec{\Sigma}_g$, effectively describing the location and the range of cluster influence. When working in the high dimensional spaces, it suits well to reduced these input by separable Gaussian, with diagonal matrix of single variances in each dimension, i.e. $\vec{\Sigma}_g = $ diag $(\sigma_{g,1}, ..., \sigma_{g,d})$.
	\vskip 0.03in
	~\\   
	{\bf Output terms:} The density $p(y|\vec{x})$ is the conditional density of the outputs $y$ given the inputs $\vec{x}$ around the cluster $g$. The presence of the conditional distribution allows the input vector $\vec{x}$ to relate with target variable $y$. In general, they are chosen to as Gaussian densities
	\begin{equation}
		\textit{p}(\textbf{y}|\textbf{x}, z_g)\sim\mathcal{N}(f(\textbf{x}, \mathbf{\beta}_g), \sigma_g)
		\label{equ:3.15}
	\end{equation}
	\begin{equation}
		p_g(y|\vec{x}) = (2\pi \sigma_g^2)^{-1/2}\exp(-\frac{1}{2}[y-f(\vec{x},\vec{\beta}_g)]^2/\sigma_g^2)
		\label{equ:3.16}
	\end{equation}
	with mean $f(\textbf{x}, \mathbf{\beta}_g)$ and variances $\sigma_g^2$ describe the local models and the error around the cluster $g$. The vector $\vec{\beta}_g$ denote the coefficient of the local model or the weight of contribution associated with the input vector $\vec{x}$.
	The $p_g(y|\vec{x})$ are normalized thus
	\begin{equation}
		\int p_g(y|\vec{x}) dy = 1 \hspace{.9cm}\forall g, \vec{x.}
		\label{equ:3.17}
	\end{equation}
	The cluster functions are chosen based on the type of supervised learning (Regression or classification) we wish to do. It is mostly chosen as linear combination of basis functions $f_i(\mathbf{x})$.
	\vspace*{0.05in}
	\bigskip
	\par \noindent The model output of the CWM is therefore weighted averagely by the local functions $f(\mathbf{x}, \vec{\beta}_g)$. The Gaussian, which are the input densities $p_g(\mathbf{x})$, controls the behavior of the local functions.
	The real problem is to find a good parameter values for
	\begin{itemize}[label = {$\bullet$}]
		\item the weights $\pi_g$,
		\item the means $\vec{\mu}_g$ and the variances $\vec{\Sigma}^2_g$ of the input density,
		\item the variances of the output terms $\sigma^2_g$,
		\item the parameters of the local functions $\vec{\beta}_g$.
	\end{itemize}
	\subsection{EM algorithm applied to CWMs }
	In the case of CWMs, as described above in section (\ref{sec:3.4}) the likelihood becomes easier by introducing a pseudo variable called a latent variable which we can interpret as unobserved data. This unobserved random variable can be imagined as sampling each pair $(x_i,y_i)$ from a single cluster with some probability.
	\par Let $Z_i \in \{1,...,G\}$ be the label of the cluster that gave rise to $(x_i,y_i)$. This random variable is unobserved. The cluster weights $\pi_g$ equation (\ref{equ:3.11}) are interpreted as the probability that $Z_i = g$ for all $g = 1 ,..., G$, implying that $Z_i$ are distributed as multinomial distribution parameterized by the cluster weights $\pi_g$. Handling cluster model parameters through maximum likelihood would be straight forward if the cluster which generates each sample was known a priori. For example, each cluster center would be the cluster mean of all points from each label. However, since this information is hidden, estimating through maximum likelihood become a nonlinear optimization problem which comes with difficulty. For this problem, EM algorithm is elegant and efficient algorithm when involving latent variables.
	\par \noindent The realization of the $Z_i$ are written as an indicator vectors $\vec{z}_i = ( z_{i1},...,z_{iG})^T$, where 
	\begin{equation}
		z_{ik} = 
		\begin{dcases}
			1, & if \hspace*{.1in}   (\mathbf{x}_i, y_i) \in \vec{z}_{ik}, \\ 
			0, & \text{otherwise}.
		\end{dcases}
		\label{equ:3.18}
	\end{equation}
	The training set is written as $\Omega_C = \{(\vec{x}_1, y_1,z_1),...,(\vec{x}_N, y_N,z_N)\}$ 
	and the complete log-likelihood $\mathcal{L}_c$ as 
	\begin{equation}
		\mathcal{L}_c(\Omega_C|\vec{\Theta}) = \sum_{i=1}^{N}\sum_{g=1}^{G} z_{ig} \log p_g(y_i|\mathbf{x}_i)p_g(\mathbf{x}_i)\pi_g,
		\label{equ:3.19}
	\end{equation} 
	where $\vec{\Theta}$ denotes the entire parameter space of the cluster weighted model, namely the weights, the means $\vec{\mu}_g$ and the variances $\vec{\sigma_g}$ of the cluster centers as well as the parameters $\vec{\beta}_g,\vec{\Sigma}_g$ of the local models.
	\par \noindent The EM algorithm optimization is initialized by the estimates $\vec{\Theta}^{(0)}$ of these parameters. One possibility, which was also used in the following, is to initialize the cluster weights uniformly i.e. $\pi_g = 1/G$, random cluster mean $\vec{\mu}_g$ by random numbers or simply picking randomly from the training data and all variances $\vec{\sigma}_g$ start with identity matrix. 
	\par \noindent In the expectation step (E-step) of the algorithm as described in chapter (\ref{c2}), the conditional expectation of $\mathcal{L}_c$ is computed with respect to the current parameter estimate, given rise to the following $Q-$ function:
	\begin{equation}
		Q(\vec{\Theta};\hat{\vec{\Theta}}) = E_\Theta\{\mathcal{L}_c(\Omega_c|\vec{\Theta})|\mathbf{x},y\}
		\label{equ:3.20}
	\end{equation} 
	The conditional expectation affects only $z_{ig}$ since the terms in the logarithm depend on $\mathbf{x}_i$ and $y_i$.
	\par \noindent E-step is effectively reduced to a calculation of the expectation of $z_{ig}$, given the observed training data. We introduce 
	\begin{equation}
		q(\mathbf{x},y;\vec{\Theta}) = E_{\vec{\Theta}}(z_{ig}|\mathbf{x},y) = p(Z_i = g|\mathbf{x}_i,y_i)
		\label{equ:3.21}
	\end{equation}
	According to the definitions from section (\ref{sec:3.4}), the posterior probability is in general given by 
	\begin{equation}
		q(\mathbf{x},y;\hat{{\vec{\Theta}}}) = \frac{p_g(y|\mathbf{x})p_g(\mathbf{x})\Theta_g}{\sum_{j=1}^{G}p_j(y,\mathbf{x})} = \frac{p_g(y|\mathbf{x})p_g(\mathbf{x})\Theta_g}{\sum_{j=1}^{G}p_j(y|\mathbf{x})p_j(\mathbf{x})\Theta_g},
		\label{equ:3.22}
	\end{equation} 
	Each cluster is able to relate with each data point through this distribution. Looking at Equation (\ref{equ:3.22}), one can see that posterior is the ratio of one cluster to all the the cluster. Given the expectation value, the $Q-$function is given by
	\begin{equation}
		Q(\vec{\Theta};\hat{\vec{\Theta}}) = \sum_{i=1}^{N}\sum_{g=1}^{G}p_g(\mathbf{x},y;\hat{\vec{\Theta}}) \log p_g(y_i|\mathbf{x}_i)p_g(\mathbf{x}_i)\pi_g
		\label{equ:3.23}
	\end{equation} 
	\par \noindent In the maximization step (M-step), the next parameter estimate $\hat{\vec{\Theta}}$ is obtained by the global maximization of the $Q-$function with respect to $\vec{\Theta}$ over the parameter space. The derivatives with respect to the desired parameter is calculated by taking the gradient with respect to the parameter of interest and setting to zero, thus obtain a new set of parameters $\vec{\Theta}$ as a function of the old parameters $\hat{\vec{\Theta}}$. This procedure is repeated until convergence.
	\par \noindent Applying the logarithmic law, $Q-$function can be decomposed as follows:
	$$Q(\vec{\Theta};\hat{\vec{\Theta}}) = \sum_{i=1}^{N}\sum_{g=1}^{G}p_g(\mathbf{x},y;\hat{\vec{\Theta}})\log p_g(y_i|\mathbf{x}_i)$$
	$$+ \sum_{i=1}^{N}\sum_{g=1}^{G}p_g(\mathbf{x},y;\hat{\vec{\Theta}})\log p_g(\mathbf{x}_i)$$
	\begin{equation}
		+ \sum_{i=1}^{N}\sum_{g=1}^{G}p_g(\mathbf{x},y;\hat{\vec{\Theta}})\log \pi_g
		\label{equ:3.24}
	\end{equation}
	This decomposition is useful as taking the gradient with respect to the parameter of interest becomes convenient. For example, the cluster weights $\pi_g$, can be computed independently of the other while others summands without the parameter of interest becomes zero automatically. Since the weights are with constraints $\sum \pi_g = 1$ and $0 \leq \pi_g \leq 1$, Lagrange multiplier is introduced as follows:
	\begin{equation}
		\frac{\partial}{\partial \pi_g} \bigg[Q(\Theta;\hat{\vec{\Theta}}) + \lambda (1 - \sum_{g=1}^{G}\pi_g)\bigg] = 0,
		\label{equ:3.25}
	\end{equation}
	\begin{figure}[H]
		\centering
		\includegraphics[width = 4.5in, height=1.5in]{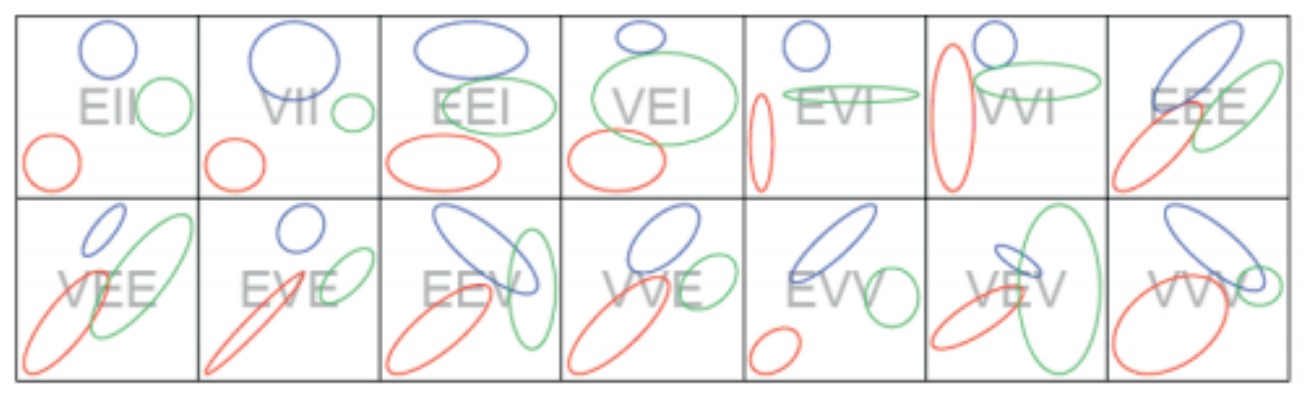}
		\centering\caption[Models Used in CWMs clustering: Example of contours of the bivariates normal component densities for the $14$ parameterization of the covariance matrix]{Models Used in  CWMs clustering: Example of contours of the bivariates normal component densities for the $14$ parameterization of the covariance matrix. Source: \cite{bouveyronetal2019}}
		\label{fig:3.1}
	\end{figure}
	\vspace{0.05in}
	\bigskip
	\par \noindent which now leads to 
	$$\pi_g = \frac{1}{N}\sum_{i=1}^{N}p_g(y_i,\mathbf{x}_i;\hat{\vec{\Theta}})$$
	which can equally be interpreted as $\sum_iz_{ig}/N$, where the unknown labels are substituted by their expectation value.
	\par \noindent The update estimates for the  means and variances of the clusters $(\vec{\mu}_g,\vec{\sigma}_g)$ are also derived by maximized $Q(\vec{\Theta};\hat{\vec{\Theta}})$. Thus, the updated means are given by 
	\begin{equation}
		\hat{\vec{\mu}}_g = \frac{\sum_{i=1}^{N}\mathbf{x}_ip_g(y_i,\mathbf{x}_i;\hat{\vec{\Theta}})}{\sum_{i=1}^{N}p_g(y_i,\mathbf{x}_i;\hat{\vec{\Theta}})}
		\label{3.26}
	\end{equation}
	\subsection{Geometrically Constrained CWMs}
	The full multivariate Gaussian for CWMs discussed above has posed a lot of problem in the estimation process. Some of the  problems which are due to high-dimensional space or large $d$ can be associated to the problem of matrix inversion caused by singularity, degeneracies of the algorithm. For full covariance matrix the parameters to be estimated are $(G-1) + Gd + G[d(d+1)/2]$. This parameters are quite a large number number. For example in the Epileptic Seizure data, with $d = 178$ and $G = 5$, this is $128,879$ parameters to be estimated, which is too large for any clustering model. Such a large numbers of parameters can lead to difficulties in estimation, including lack of precision or even cause the algorithm to degenerate. They also reduce the computational speed of the algorithms. In order to mitigate this problem, \cite{BanfieldandRaftery1993} and \cite{CeleuxandGovaert1995a} introduced the eigenvalue decomposition of the cluster covariance matrix $\Sigma_g$, in the form
	\begin{equation}
		\Sigma_g = \lambda_g D_g A_g D_g^T.\label{equ:3.27}
	\end{equation} 
	In Equation (\ref{equ:3.27}), $D_g$ is the matrix of the eigenvectors of $\Sigma_g$, $A_g = diag\{A_{1,g},...,A_{d,g}\}$ is a diagonal matrix whose elements are proportional to the eigenvalues of $\Sigma_g$ arranged in a descending order, and $\lambda_g$ is the constant associated with the proportionality.
	\par \noindent Each elements in this decomposition corresponds to a particular geometric property of the $g$th component. The matrix of the eigenvectors $D_g$ determines its orientation in $\mathrm{R}^d$. The diagonal matrix of scaled eigenvalues $A_g$ governs its shape. The region where the $g$th is densely concentrated can be determined by the maximum number of the shape in the plane. For example, if $A_{1,g} \gg A_{2,g}$, then the $g$th component is tightly concentrated around a line in $\mathrm{R}^d$. If $A_{1,g} \approx A_{2,g} \gg A_{3,g}$, then the $g$th component is concentrated in a two-dimensional plane in $\mathrm{R}_d$. If all the values of $A_{j,g}$ are approximately equal, then the $g$th component is roughly equal. The constant of proportionality determines the volume. This is proportional to $\lambda_g^d|A_g|$ where $||A_g$ is determinant of $A_g$ preferably constrained to be equal to $1$. Parsimony occurs in different ways using the decomposition by either constraining any or all of the volume, shape or orientation to be to be equal or varied across the clusters. Also, the covariance matrix can be forced to be spherical i.e. Identity matrix $I$. Whenever the covariance matrix is spherical, there are two univariate models, and $14$ possible models in multivariate case. Figure (\ref{fig:3.1}) shows the examples of contours of the component densities for the various models in the two-dimensional case with two mixture components.
	\vspace{0.05in}
	\begin{table}[H]
		\centering\caption[Parameterizations of the covariance matrix through Eigenvalue decomposition]{Parameterizations of the covariance matrix $\Sigma_g$ through Eigenvalue decomposition. $A$ denotes a diagonal matrix}\vspace{0.4cm}
		\label{tab:3.1}
		\begin{tabular}{c@{\hspace*{.3in}}c@{\hspace*{.3in}}c@{\hspace*{.3in}}c@{\hspace*{.3in}}c@{\hspace*{.3in}}c@{\hspace*{.3in}}}
			\hline
			Identifier&Model&Distribution&Volume&Shape&Orientation\\ 
			\hline
			E&$-$&Univariate&Equal&Not required&Not required\\
			V&$-$&Univariate&Variable&Not required&Not required\\
			\vspace*{0.05in}
			\\
			\hline	
			EII&$\lambda I$&Spherical&Equal&Equal&Not required\\
			VII&$\lambda_g I$&Spherical&Variable&Equal&Not required\\
			\vspace*{0.05in}
			\\
			\hline
			EEI&$\lambda A$&Diagonal&Equal&Equal&Axis-aligned\\
			VEI&$\lambda_g A$&Diagonal&Variable&Equal&Axis-aligned\\
			EVI&$\lambda A_g$&Diagonal&Equal&Variable&Axis-aligned\\
			VVI&$\lambda_g A_g $&Diagonal&Variable&Variable&Axis-aligned\\
			\vspace*{0.05in}
			\\
			\hline
			EEE&$\Sigma$&Ellipsoidal&Equal&Equal&Equal\\
			VEE&$\lambda_g D A D^T$&Ellipsoidal&Variable&Equal&Equal\\
			EVE&$\lambda D A_g D^T$&Ellipsoidal&Equal&Variable&Equal\\
			EEV&$\lambda D_g A D_g^T$&Ellipsoidal&Equal&Equal&Variable\\
			\vspace*{0.05in}
			\\
			\hline
			VVE&$\lambda_g D A_g D^T$&Ellipsoidal&Variable&Variable&Equal\\
			VEV&$\lambda_g D_g A D_g^T$&Ellipsoidal&Variable&Equal&Variable\\
			EVV&$\lambda D_g A_g D_g^T$&Ellipsoidal&Equal&Variable&Variable\\
			VVV&$\Sigma_g$&Ellipsoidal&Variable&Variable&Variable\\
			\vspace*{0.05in}
			\\
			\hline
		\end{tabular}
	\end{table}
	\bigskip
	\par \noindent These constrained models can have extremely fewer parameters that need to be estimated independently than the full covariance model, while fitting the sample data almost as well. The constrained models can yield more precise estimates of model parameters, accurate out-of-sample predictions, and easy interpretability of parameter estimates. Moreover, the model have $Gd$ parameters for the component means $\vec{\mu}_g$, and $(G-1)$ parameters for the mixture proportions $\pi_g$. 
	\par Table (\ref{tab:3.1}) shows the multivariate models denoted by three-letter identifier where "E" stands for Equal and "V" stands for variable. If the first letter is "E" it means the volume is equal/constant across the clusters, and "V" if varied across. In the same vein, the second letter "E" represents equal shape and "V" if not, so that for all $g = 1,...,G$, the shape matrices $A_g \equiv A$. "I" stands for spherical when the $A_g = I$ for $g = 1,...,G$. Finally, if "E" is located at the the third position, then the $D_g$ of eigenvectors specify the cluster orientations are equal $D_g \equiv D$ for $g = 1,..., G$, "V" if they are not constrained, and "I" if the clusters are spherical such that $D_g = I$ for $g = 1,.., G$. 
	\par Table (\ref{tab:3.2}) shows the numbers of parameters needed to specify the covariance matrix for each model in the $178$-dimensional five-component case, $d = 178, G = 5$, three-dimensional five-component case, $d = 3, G = 5$ gotten from the Epileptic seizure recognition data before dimensionality reduction, and after dimensionality reduction, respectively. Before performing the dimensionality reduction, we note that CWM is impracticable. These results are obtained by noting that for one mixture component, the volume is specified by $1$ parameter, the shape by $(d-1)$ parameters, and the orientation by $d(d-1)/2$. The potential gain in the combination of parsimony and dimensional reduction is far higher than the gain achieved from only parsimony compared to the full covariance matrix parameters. In the most extreme case in Table (\ref{tab:3.2}), in the $178$-dimensional case with $5$ mixture components, the VVV model requires $79,655$ parameters to represent the covariance matrices, whereas  the same VVV requires $30$ parameters with the combination of dimensionality reduction and eigenvalue decomposition. Although, there are some gains in parsimony, however it has been observed that the most parsimonious models do not always fit the data adequately. Moreover, the number of parameters to be estimated in parsimonious model is still outrageously high, and the preferable solution would be to apply some steps further parsimonious method to the results of the parsimonious model. However, this might not achievable if the computational time is a priority. 
	\vspace{0.05in}
	\begin{table}[H]
		\centering\caption[Numbers of parameters needed to specify the covariance matrix]{Numbers of the parameters needed to specify the covariance matrix for \\ \centering models used CWMs and CWMs-tSNE}\vspace{0.4cm}
		\label{tab:3.2}
		\begin{tabular}{c@{\hspace*{.3in}}c@{\hspace*{.3in}}c@{\hspace*{.3in}}c@{\hspace*{.3in}}}
			\hline
			Model&General&$d = 3, G = 5 $&$d = 178, G = 5$\\ 
			\hline
			E&$-$&$-$&$-$\\
			V&$-$&$-$&$-$\\
			\vspace*{0.05in}
			\\
			\hline	
			EII&$1$&1&1\\
			VII&$G$&$5$&5\\
			\vspace*{0.05in}
			\\
			\hline
			EEI&$d$&3&178\\
			VEI&$G + (d-1)$&7&182\\
			EVI&$1 + G(d-1)$&11&886\\
			VVI&$Gd$&15&890\\
			\vspace*{0.05in}
			\\
			\hline
			EEE&$d(d+1)/2$&6&15931\\
			VEE&$G+(d+2)(d-1)/2$&10&15935\\
			EVE&$1+(d+2G)(d-1)/2$&14&16639\\
			EEV&$1+(d-1)+G[d(d-1)/2]$&18&78943\\
			\vspace*{0.05in}
			\\
			\hline
			VVE&$G+(d+2G)(d-1)/2$&18&16643\\
			VEV&$G+(d-1)+G[d(d-1)/2]$&22&78947\\
			EVV&$1+G(d+2)(d-1)/2$&26&79651\\
			VVV&$G[d(d+1)/2]$&30&79655\\
			\vspace*{0.05in}
			\\
			\hline
		\end{tabular}
	\end{table}
	\vspace*{0.1in}
	~\\~
	Alternatively, the best solution would be to perform dimensionality reduction before using parsimony. Unfortunately, Eigenvalue decomposition method does what we can call a "local parameter reduction" when the "global feature" remains huge. Fitting the huge original high-dimensional data irrespective of the parsimony encumbers CWMs model. Consequentially, reducing the classification power, slows the computation speed, and lead to misinterpretation of the result. This becomes a challenge in CWMs models.
		\begin{figure}[H]
		\centering
		\begin{minipage}[b]{0.45\textwidth}
			\includegraphics[width = 3in, height=2in]{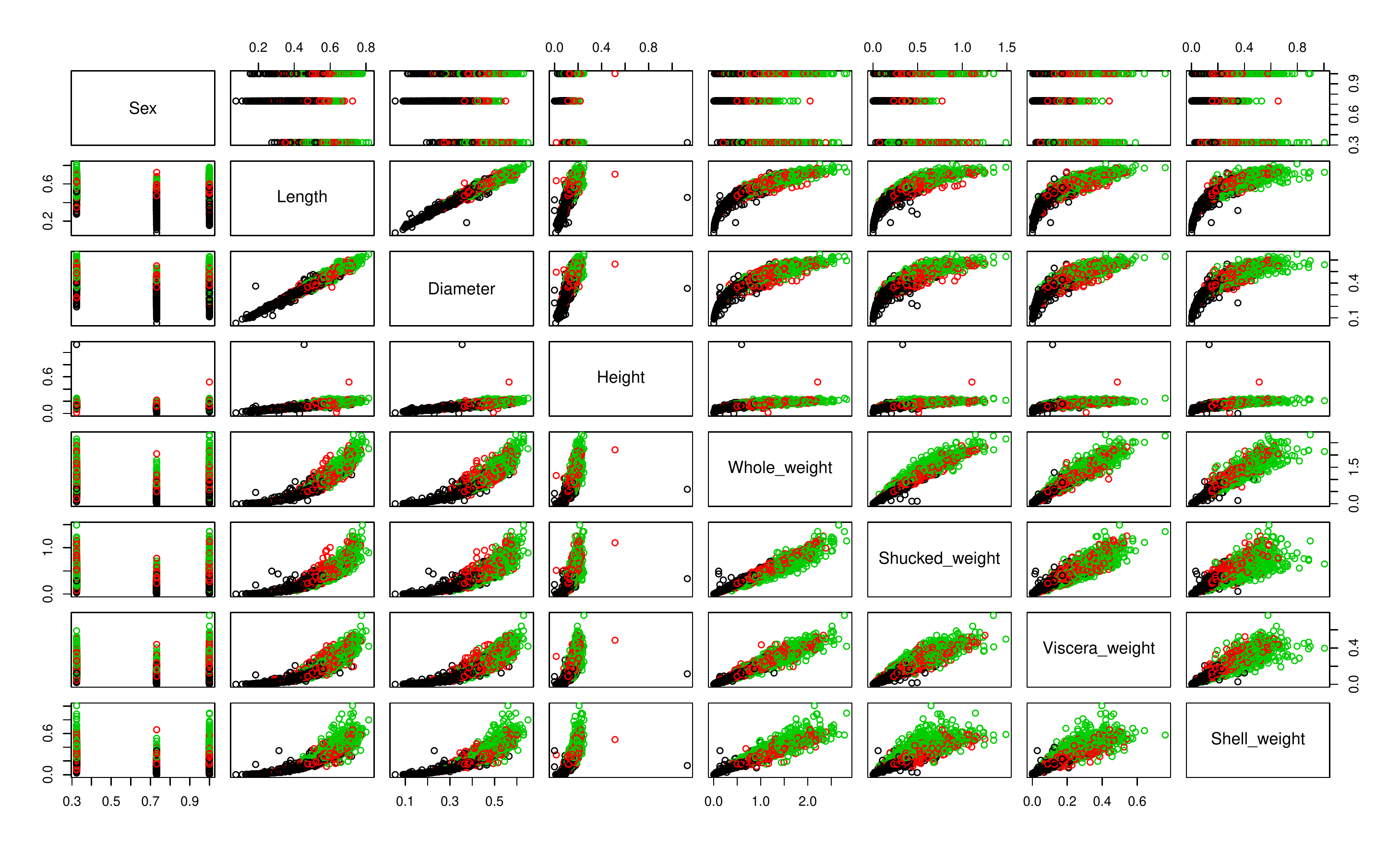}
			\centering\caption[The visualization of the descriptive summary of the original Abalone data]{The visualization of the descriptive summary of the original Abalone data colored according to the grouping of the Rings: Black ($1-8$), Red ($9-10$), and Green ($\geq 11$)}
			\label{fig:3.2}
		\end{minipage}
		\hfil
		\begin{minipage}[b]{0.45\textwidth}
			\includegraphics[width = 3in, height=2in]{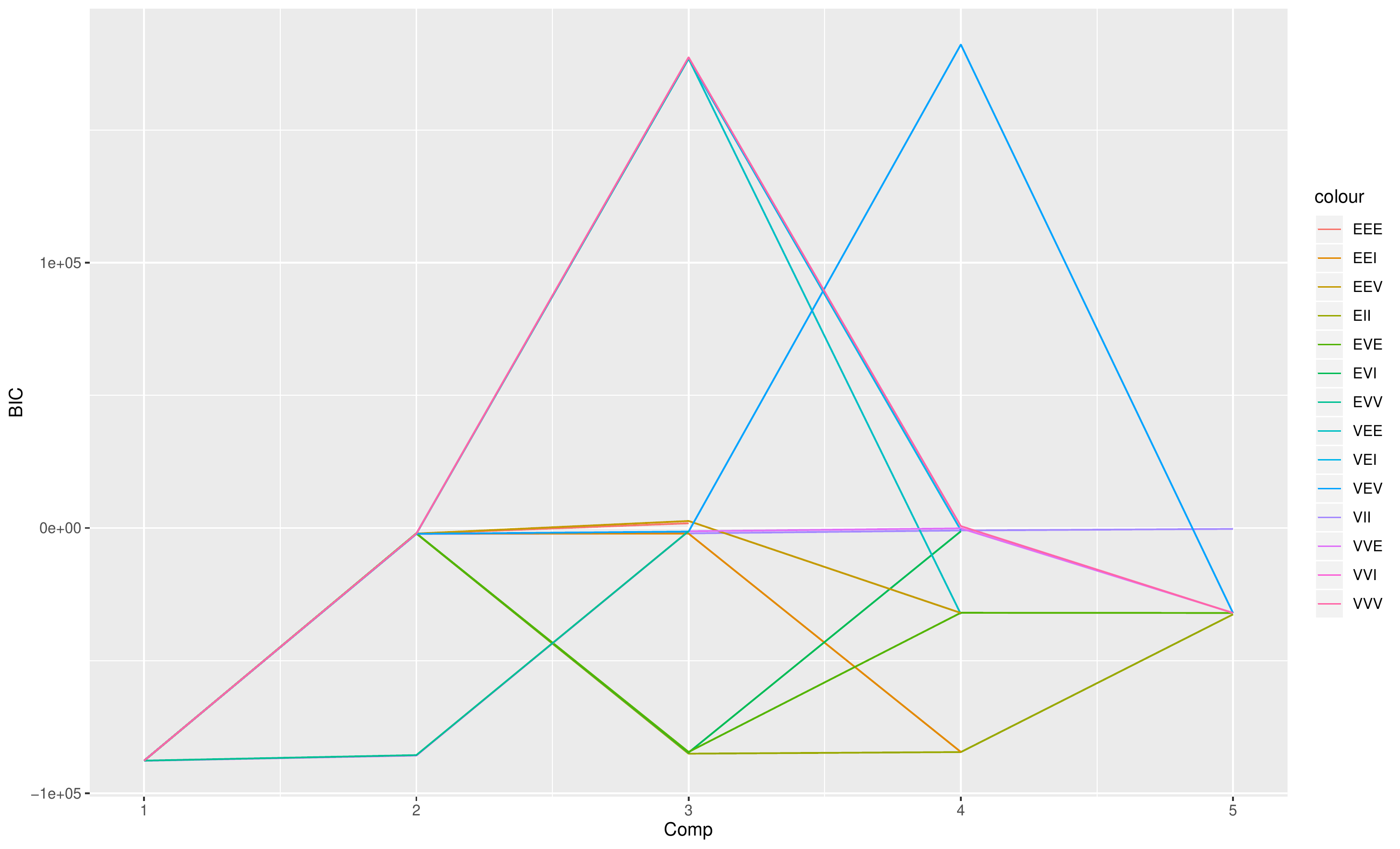}
			\centering\caption[Model selection for the Abalone data using BIC values of the fourteen models]{Model selection for the Abalone data using BIC values of the fourteen models. The BIC produced by three models select the correct number of components.} 
			\label{fig:3.3}
		\end{minipage}
	\end{figure}
	\subsection{Dimensionality reduction}
	Given the high-dimensional nature of the dataset considered here, a preprocessing step of feature extraction is of great importance to reduce the computational burden and time complexity before fitting the CWMs model. The considered preprocessing step proceeds as follows; first of, we fit the feature set to the tSNE, and afterwards we project the test unit nonlinearly to obtain the low-dimensional subspace. Without dimensionality reduction process, CWMs can be so limited by the high-dimensional data which slows down the computational speed and hamper the clustering performance of the model. The subspace of the original features are then filtered into the CWMs for clustering analysis. Moreover, the visualization of the high-dimensional data is made possible by tSNE technique. Here, we present both low-dimensional data and high-dimensional data with features running to the order of hundred.	
	\section{Results and Discussion}
	This section illustrates some real data applications of the linear CWMs defined above with a substantive high dimensionality. The analysis is performed using the $\mathbf{R}$ package for CWMs called $\mathbf{FlexCWM}$, [\cite{mazzaetal2018}].  
	\subsection{Abalone data}
	The first application concerns the prediction of age of abalone from physical measurements. The data was taken from UCI database with the original sources of Marine Resources Division and \cite{samwaugh1995}; \cite{warwicketal1994}. The age of abalone was determined by counting the number of rings through a microscope after cutting the shell through the cone, and staining it. The analysis presented below uses all the variables in the dataset. The following are the attributes of the data;
	Sex: Male (M), Female (F), and Infant (I), Length: Longest shell measurement, Diameter: Perpendicular to length, Height: With meat in shell, Whole.Weight: Whole abalone, Shucked.Weight: Weight of meat, Viscera.Weight: Gut weight after bleeding, Shell.Weight: Gut weight after being dried, Rings: Age in years of the Abalone.
	\par \noindent There are $G = 3$ groups of abalone with respect to Sex variables: $M = 1528$, $F = 1307$, and $I = 1342$. First off, we use the whole variables and check the effect on the clustering power of the linear CWM. 
	\par \noindent We compare the Bayesian Information criteria (BIC) produced by fourteen different parsimonious models.
	\par \noindent Figure (\ref{fig:3.2}) concerns the observed labeled data. This graphical representation is the visualization of the descriptive summary of the abalone data. The observations are color-coded according to the group of the Rings variable grouped into $3$- class category; $1-8$, $9-10$, and $\geq 11$. The goal is to classify the abalone according to their age group. The nonlinear projection from the original feature space to low dimensional feature space is performed. However, the goal is not to separate the observation to their respective classes but to reduce the dimension of the data which leads to the removal of any multi-collineariy among the 
	\begin{figure}[h]
		\centering
		\begin{minipage}[b]{0.45\textwidth}
			\includegraphics[width = 3in, height=2in]{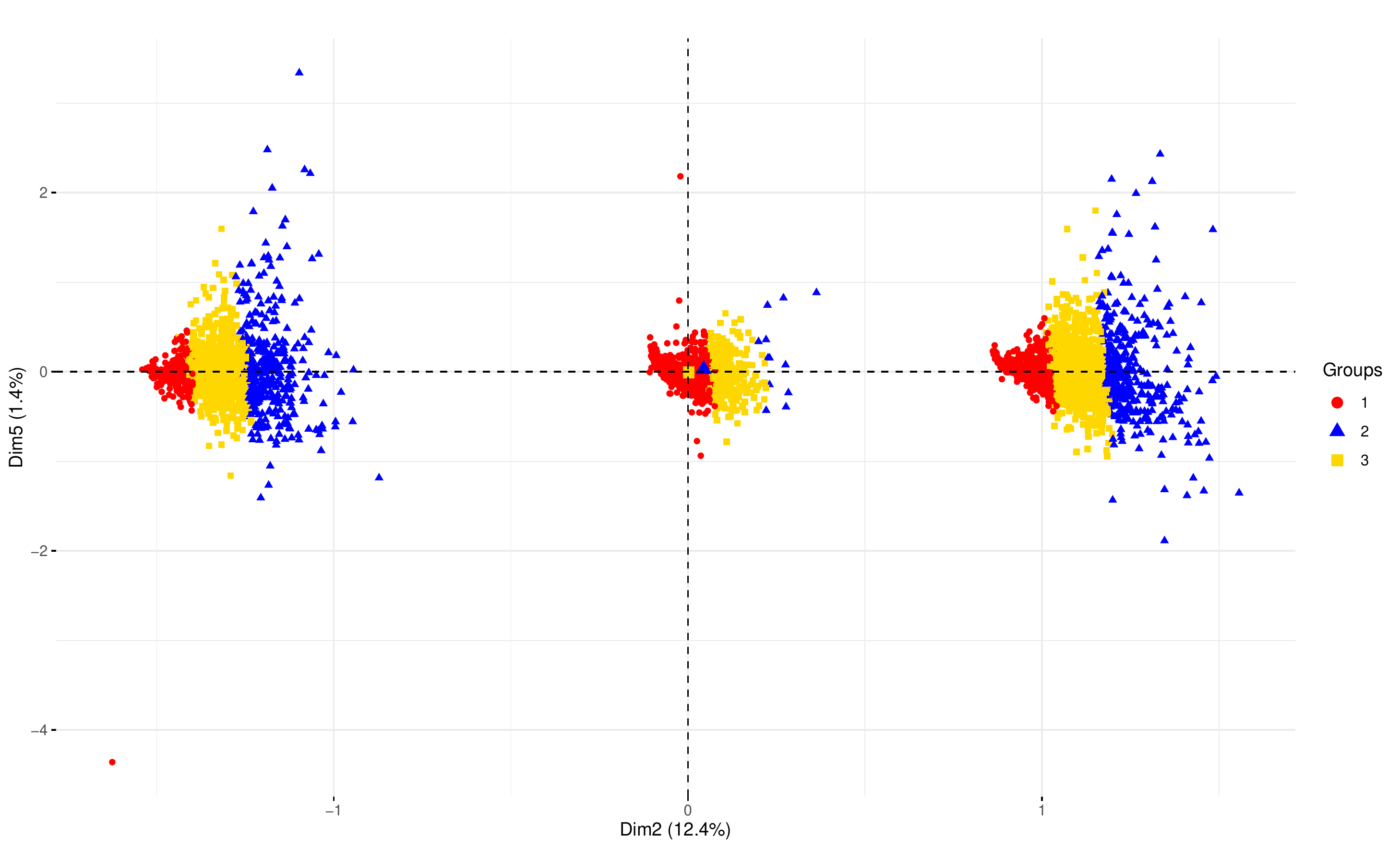}
			\centering\caption[The classification plot of CWM-tSNE for $G = 3$]{The classification plot of CWM-tSNE for $G = 3$ with model VVV selected by BIC.}
			\label{fig:3.4}
		\end{minipage}
		\hfil
		\begin{minipage}[b]{0.45\textwidth}
			\includegraphics[width = 3in, height=2in]{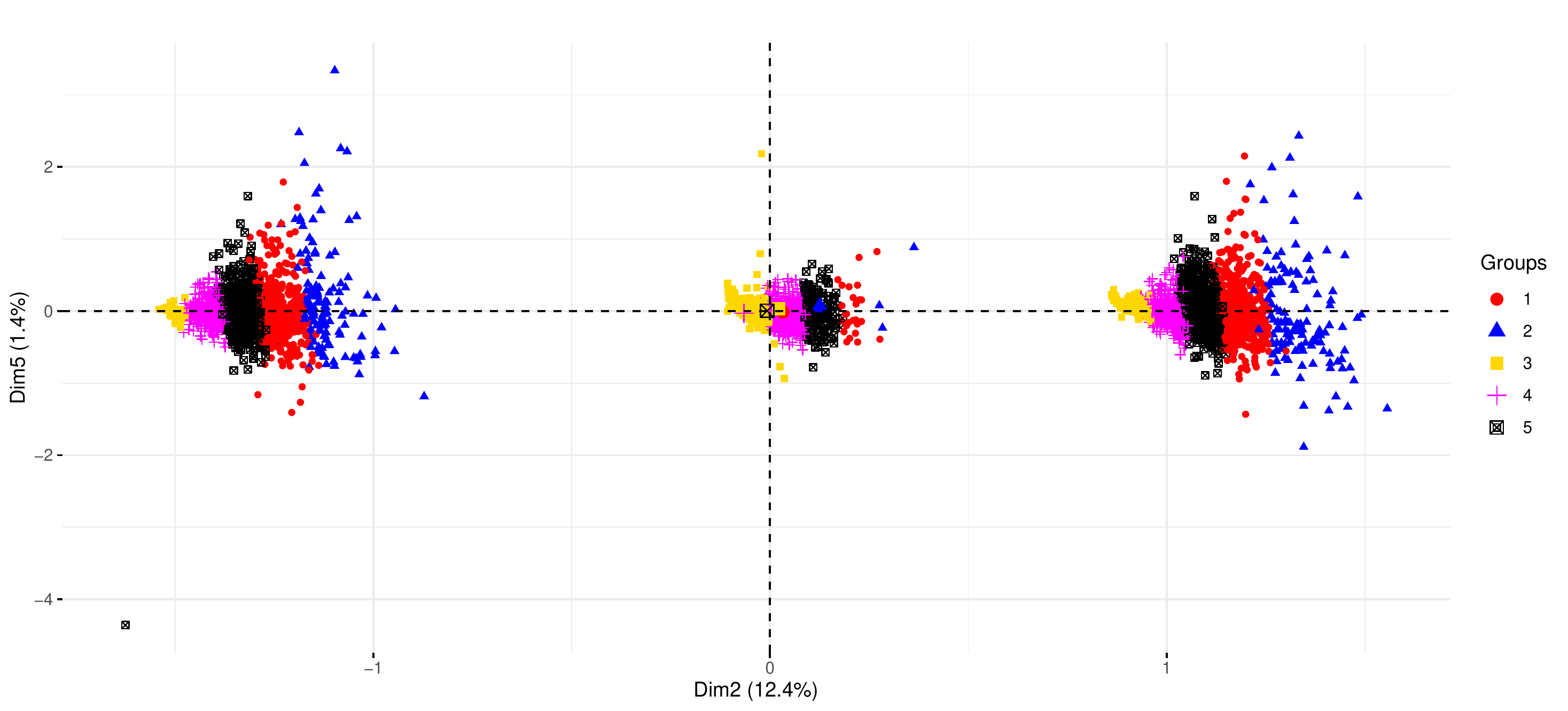}
			\centering\caption[The classification plot of CWM-tSNE for $G = 4$]{The classification plot of CWM-tSNE for $G = 4$ as suggested by the BIC} 
			\label{fig:3.5}
		\end{minipage}
	\end{figure}
	\vspace*{0.01in}
	\bigskip
	\begin{table}[H]
		\centering\caption[The selection of the best model among 14 models according to the BIC]{The selection of the best model among 14 models according to the BIC is VVV}\vspace{0.4cm}
		\label{tab:3.3}
		\begin{tabular}{r@{\hspace*{.3in}}r@{\hspace*{.3in}}r@{\hspace*{.3in}}r@{\hspace*{.3in}}r@{\hspace*{.3in}}r@{\hspace*{.3in}}}
			\hline
			Model&comp1&comp2&comp3&comp4&comp5\\ 
			\hline
			EII&-87696.1&-$\vec{2109.2}$&-85062.3&-84460.1&-32541.1\\
			VII&-87696.1&-2230.8&-2024.8&-917.4&-$\vec{341.9}$\\
			EEI&-87688.0&-2115.9&-$\vec{2101.1}$&-84412.6&Not Estimated\\
			VEI&-87688.0&-2184.1&$\vec{176940.8}$&-838.7&Not Estimated\\
			EVI&-87688.0&-2006.1&-84632.3&-$\vec{1225.5}$&Not Estimated\\
			VVI&-87688.0&-2085.3&$\vec{177254.5}$&Not Estimated&-32220.5\\
			EEE&-87687.8&-2122.8&$\vec{1793.2}$&Not Estimated&-32554.9\\
			VEE&-87687.8&-2146.5&$\vec{176932.7}$&-32459.0&Not Estimated\\
			EVE&-87712.7&-$\vec{1999.1}$&-84415.3&-31949.4&-32016.0\\
			EEV&-87687.8&-2086.2&$\vec{2651.1}$&-32021.3&Not Estimated\\
			VVE&-87734.3&-85777.6&-1219.1&-$\vec{118.8}$&-32002.6\\
			VEV&-87687.8&-2186.5&-1337.2&$\vec{182260.9}$&-32072.1\\
			EVV&-87687.8&-85606.0&-1196.0&Not Estimated&-$\vec{1063.1}$\\
			VVV&-87687.8&-2108.3&$\vec{177462.9}$&715&-32028.74\\
			\hline
		\end{tabular}
	\end{table}
	\vspace*{0.05in}
	\bigskip
	\noindent features. Afterwards, we filtered the projected feature into CWMs. This is always better in terms of speed and accuracy. We perform the analysis on the original data and the parsimonious models selected the same number of component as the projected data. This assures us that the low-dimensional data is a good representation of the original data. However, all the eight information criteria have an extremely high number produced by the original data. This might be due to redundancy in the feature of the original data. 
	\par The data can be seen as a nested cluster or as having both global and local components, i.e. cluster through the sex variable of the Abalone which are male (M), Infant (I), and Female (F), and the grouping through the age of the Abalone. This makes the data extreme difficult to separate. The previous work by \cite{samwaugh1995} also confirmed the presence of overlap in the data while he suggested additional information to separate the class completely using the affine combinations.  We note that it is easier to separate the data with respect to the sex variable while the age group remains cluttered together. This can hinder the performance of the clustering algorithm. 
	\par \noindent Figure (\ref{fig:3.3}) shows the values of BIC for the models in the CWMs-tSNE with G ranging from $1,...,5$. We show the plot resulting from BIC. In CWMs-tSNE model, the four models that provide the largest values for the BIC were VEI, VVI, VEE, VVV with values: $176940.8, 177254.5,176932.7$, and $177462.9$. In Table (\ref{tab:3.3}), we presented only the BIC values for the 14 models considered because the eight information criteria agreed in selecting the same number of components. The best models are distinguished with boldface.
	\begin{table}[H]
		\centering\caption[Adjusted Rand Index and its variants of the three-component Model for Abalone data]{Adjustment Rand Index and its variants of the three-component Model for Abalone data. According to the BIC, the models VEI, VVI, VEE, and VVV give $\text{ARI} = 1$}\vspace{0.4cm}
		\label{tab:3.4}
		\begin{tabular}{r@{\hspace*{.3in}}r@{\hspace*{.3in}}r@{\hspace*{.3in}}r@{\hspace*{.3in}}r@{\hspace*{.3in}}r@{\hspace*{.3in}}}
			\hline
			Model&Rand&HA&MA&FM&Jaccard\\ 
			\hline
			EII&0.780&0.576&0.576&0.777&0.603\\
			VII&$0.782$&$0.471$&$0.471$&$0.627$&$0.447$\\
			EEI&$0.822$&$0.599$&$0.599$&$0.733$&$0.577$\\
			VEI&$\vec{1.000}$&$\vec{1.000}$&$\vec{1.000}$&$\vec{1.000}$&$\vec{1.000}$\\
			EVI&$0.794$&$0.513$&$0.513$&$0.662$&$0.490$\\
			VVI&$\vec{1.000}$&$\vec{1.000}$&$\vec{1.000}$&$\vec{1.000}$&$\vec{1.000}$\\
			EEE&$0.823$&$0.603$&$0.603$&$0.735$&$0.582$\\
			VEE&$\vec{1.000}$&$\vec{1.000}$&$\vec{1.000}$&$\vec{1.000}$&$\vec{1.000}$\\
			EVE&$0.781$&$0.576$&$0.576$&$0.777$&$0.603$\\
			EEV&$0.809$&$0.571$&$0.571$&$0.715$&$0.556$\\
			VVE&0.782&0.488&0.488&0.647&0.475\\
			VEV&$0.957$&$0.901$&$0.901$&$0.934$&$0.872$\\
			EVV&$0.798$&$0.519$&$0.520$&$0.666$&$0.493$\\
			VVV&$\vec{1.000}$&$\vec{1.000}$&$\vec{1.000}$&$\vec{1.000}$&$\vec{1.000}$\\
			\hline
		\end{tabular}
	\end{table}
	
	\par \noindent Also, the ARI and its variants are presented in Table (\ref{tab:3.4}). The ARI for the models selected by the BIC as shown in Table (\ref{tab:3.3}) is $1$. In contrast, according to \cite{samwaugh1995} the Cascading-Correlation with no hidden nodes and with $5$ hidden nodes had $24.8\%$ and $26.2\%$ , while C4.5 achieved $21.5\%$, Linear Discriminant Analysis (LDA) achieved $0.0\%$, and the $k=5$ Nearest Neighbor got $3.57\%$ accuracy.
	\subsection{Protein data}
	The goal of the second application is to cluster the localization site of proteins. The protein data created by \cite{paulandkental1996} and is available in the UCI database. The data consist of seven input variables and class variable. There are $N = 336$ observations and attributes information is as follows;\\
	Sequence Name: Accession number for the SWISS-PORT database, mcg: McGeoh's method for signal sequence recognition, gvh: Von Heijne's method for signal sequence recognition, lip: von Heijne's signal Peptidase II consensus sequence score, chg: Presence of charge on N-terminus of predicted lipoproteins, aac: Score of discriminant analysis of the amino acid content of outer membrane, alm1: Score of the ALOM membrane spanning region prediction program, alm2: Score of ALOM program after excluding putative 
		\begin{figure}[H]
		\centering
		\begin{minipage}[b]{0.45\textwidth}
			\includegraphics[width = 3in, height=2in]{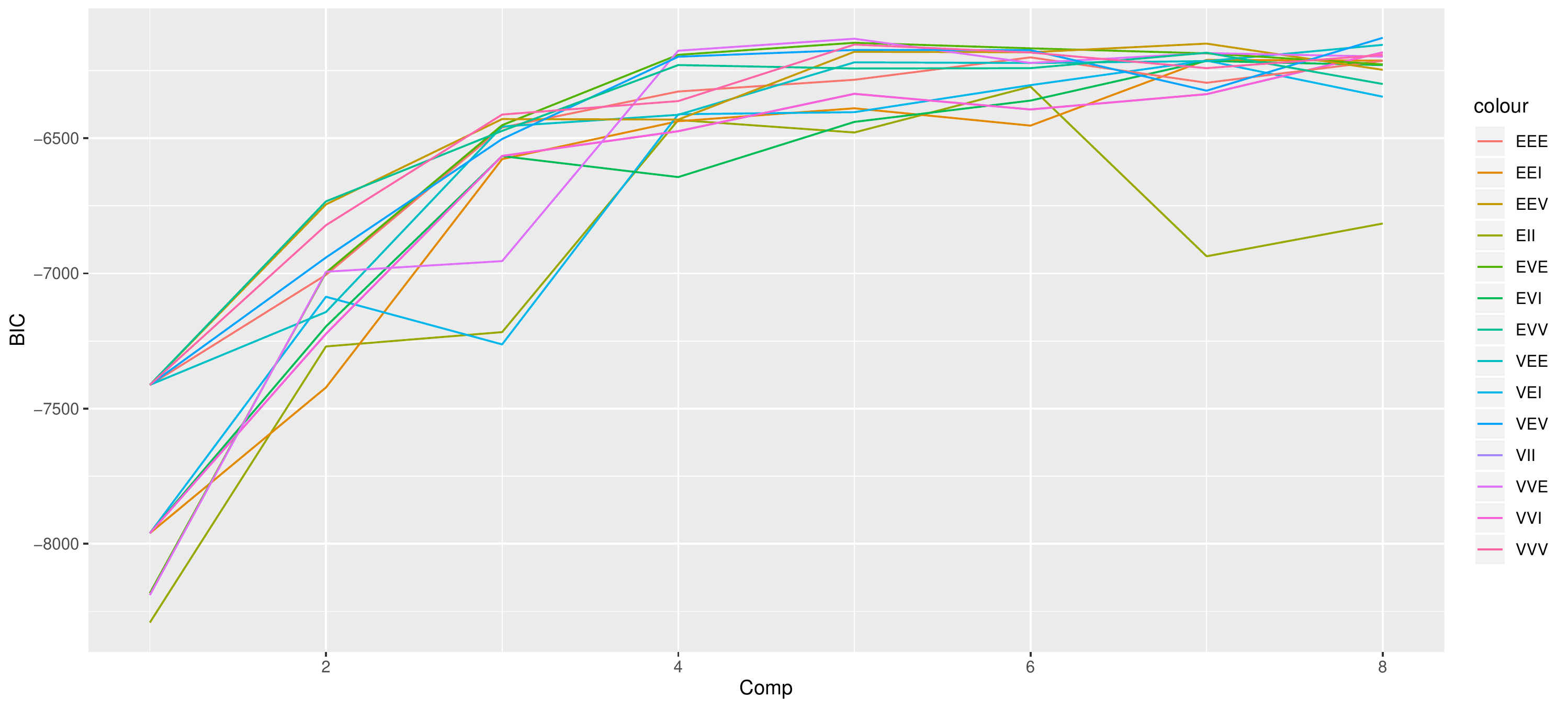}
			\centering\caption[Model selection for the protein data using BIC values of the fourteen models]{Model selection for the protein data using BIC values of the fourteen models. The BIC produced by five models select the correct number of components.}	
			\label{fig:3.6}
		\end{minipage}
		\hfil
		\begin{minipage}[b]{0.45\textwidth}
			\includegraphics[width = 3in, height=2in]{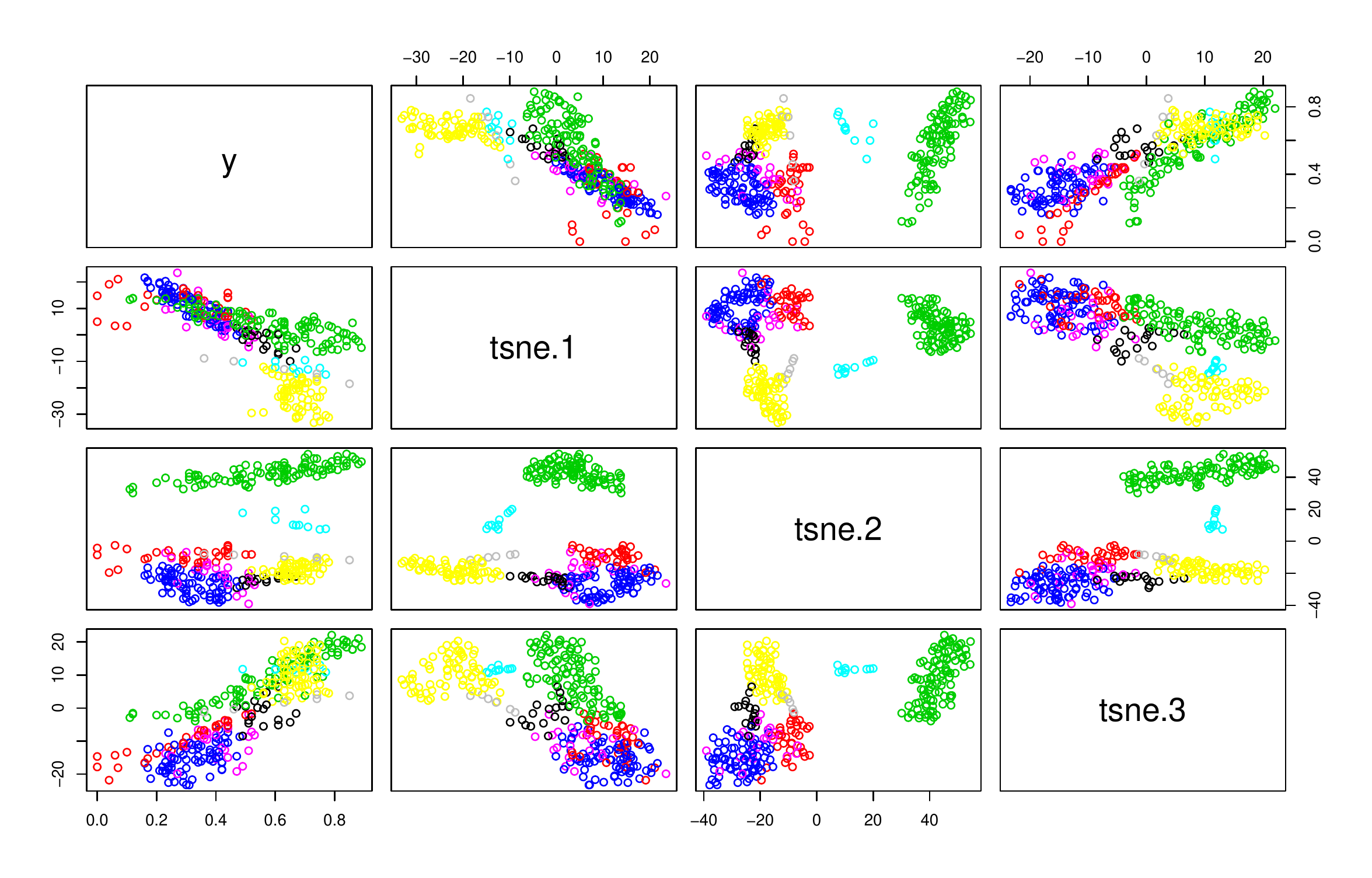}
			\centering\caption[The plot produced by CWMs after dimension reduction via tSNE]{The plot produced by CWMs after dimension reduction via tsne. CWMs selected eight components which aligns to the true class of the localization site of protein.} 
			\label{fig:3.7}
		\end{minipage}
	\end{figure}
	\vspace{0.01in}
	\bigskip
	\noindent cleavable signal regions from the sequence. According to the framework of CWMs, we transformed the multiclass response called the localized site by adding the $0.5$ and taking the logarithm of the result. This is done to transform from a categorical variable to continuous. We pretended as if the true clustering is not known apriori and check which model would perform the best among the fourteen parsimonious models. In order to visualize the BIC values, Figure (\ref{fig:3.6}) shows the BIC plot for the protein data, using the $\mathbf{R}$ commands provided by the $\mathbf{FlexCWM}$ package. Values are shown for up to $G_{max} = 8$ components and for the $14$ covariance models estimated in the same package, i.e. for $8 \times 14$ different competing models in all. BIC selects the model with six mixture components and the EEE with 10 other covariance specifications, in which all the covariance matrices are either equal or varied. However, BIC selects the five models such as VII, VVI, VEE, and VEV with eight mixture components. 
	\par \noindent Table (\ref{tab:3.5}) lists the values of the BIC for the fourteen models. The values of the BIC according to the Table (\ref{tab:3.5}) are $-6183.0$ (VII), $-6183.0$ (VVI), $-6155.1$ (VEE), and $-6129.2$ (VEV). Among all the models considered, the value of the BIC $-6309.6$ produced by EII is the worst model. Table (\ref{tab:3.6}) is generated by comparing the clusters produced by the BIC values with the true class of the localized site using the varieties of ARI. VVE model shows higher values of the ARI among all the models.
	According to the selection of the component produced by the CWM-tSNE model, Figure (\ref{fig:3.6}) shows the classification for the VVI selected model with respect to the number of cluster produced by the CWMs-tSNE. We note that AWE gives wrong number of clusters throughout the analysis. The protein data has been analyzed by \cite{paulandkental1996}. In their work of "A Probabilistic Classification System for predicting the Cellular Localization Sites of Proteins", their model achieved $81\%$ classification accuracy. Also similar accuracy has been achieved for Binary Decision Tree and Bayesian Classification methods.
	\begin{table}[H]
		\centering\caption[The comparison of the BIC produced by the fourteen models]{The comparison of the BIC produced by the fourteen parsimonious models after performing the dimensionality reduction}\vspace{0.4cm}
		\label{tab:3.5}
		\begin{tabular}{r@{\hspace*{.15in}}r@{\hspace*{.15in}}r@{\hspace*{.15in}}r@{\hspace*{.15in}}r@{\hspace*{.15in}}r@{\hspace*{.15in}}r@{\hspace*{.15in}}r@{\hspace*{.15in}}r}
			\hline
			Model&comp1&comp2&comp3&comp4&comp5&comp6&comp7&comp8\\ 
			\hline
			EII&-8291.1&-7270.3&-7217.3&-6433.0&-6479.2&-$\vec{6309.6}$&-6937.0&-6815.8\\
			VII&-7960.9&-7223.6&-6565.9&-6474.5&-6336.2&-6394.1&-6337.5&-$\vec{6183.0}$\\
			EEI&-7960.9&-7422.0&-6577.6&-6437.5&-6389.8&-6453.8&-$\vec{6210.9}$&-6213.5\\
			VEI&-7960.9&-7086.6&-7262.7&-6411.8&-6404.0&-6304.2&-$\vec{6213.3}$&-6346.9\\
			EVI&-7960.9&-7196.4&-6566.5&-6644.1&-6440.1&-6361.3&-$\vec{6214.6}$&-6230.0\\
			VVI&-7960.9&-7223.6&-6565.9&-6474.5&-6336.2&-6394.1&-6337.5&-$\vec{6183.0}$\\
			EEE&-7412.5&-7005.6&-6460.8&-6327.7&-6284.3&-$\vec{6201.5}$&-6295.4&-6214.2\\
			VEE&-7412.5&-7143.0&-6457.2&-6413.9&-6219.8&-6222.3&-6215.4&-$\vec{6155.1}$\\
			EVE&-8182.5&-6997.2&-6452.2&-6192.0&-$\vec{6147.5}$&-6168.1&-6186.7&-6227.5\\
			EEV&-7412.5&-6745.1&-6429.6&-6431.5&-6180.8&-6182.5&-$\vec{6150.6}$&-6247.5\\
			VVE&-8189.6&-6994.1&-6954.8&-6176.8&-$\vec{6132.6}$&-6222.1&-6185.7&-6198.5\\
			VEV&-7412.5&-6941.4&-6503.3&-6199.0&-6174.0&-6174.7&-6324.7&-$\vec{6129.2}$\\
			EVV&-7412.5&-6733.9&-6473.3&-6229.9&-6242.5&-6241.3&-$\vec{6184.3}$&-6299.9\\
			VVV&-7412.5&-6822.0&-6412.8&-6363.1&-$\vec{6154.2}$&-6183.6&-6241.4&-6192.4\\
			\hline
		\end{tabular}
	\end{table}
	\vspace*{0.1in}
	\begin{table}[H]
		\centering\caption[The comparison of the varieties of ARI produced by the fourteen parsimonious models]{Adjustment Rand Index and its variants of the fourteen parsimonious models to select the hidden structure or cluster in the protein data}\vspace{0.4cm}
		\label{tab:3.6}
		\begin{tabular}{r@{\hspace*{.3in}}r@{\hspace*{.3in}}r@{\hspace*{.3in}}r@{\hspace*{.3in}}r@{\hspace*{.3in}}r@{\hspace*{.3in}}}
			\hline
			Model&Rand&HA&MA&FM&Jaccard\\ 
			\hline
			EII&0.821&0.478&0.483&0.603&0.411\\
			VII&$0.794$&$0.429$&$0.435$&$0.567$&$0.389$\\
			EEI&$0.799$&$0.413$&$0.419$&$0.549$&$0.360$\\
			VEI&$0.800$&$0.428$&$0.434$&$0.562$&$0.378$\\
			EVI&$0.777$&$0.336$&$0.343$&$0.484$&$0.299$\\
			VVI&0.798&0.414&0.420&0.550&0.364\\
			EEE&$0.799$&$0.412$&$0.419$&$0.549$&$0.359$\\
			VEE&$0.848$&$0.587$&$0.591$&$0.690$&$0.522$\\
			EVE&$0.816$&$0.468$&$0.474$&$0.594$&$0.405$\\
			EEV&$0.788$&$0.409$&$0.415$&$0.551$&$0.373$\\
			VVE&$\vec{0.866}$&$\vec{0.646}$&$\vec{0.649}$&$\vec{0.737}$&$\vec{0.581}$\\
			VEV&$0.803$&$0.446$&$0.452$&$0.578$&$0.396$\\
			EVV&$0.788$&$0.406$&$0.411$&$0.547$&$0.368$\\
			VVV&$0.783$&$0.372$&$0.379$&$0.516$&$0.334$\\
			\hline
		\end{tabular}
	\end{table}
	\begin{figure}[H]
		\centering
		\begin{minipage}[b]{0.45\textwidth}
			\includegraphics[width = 3in, height=2in]{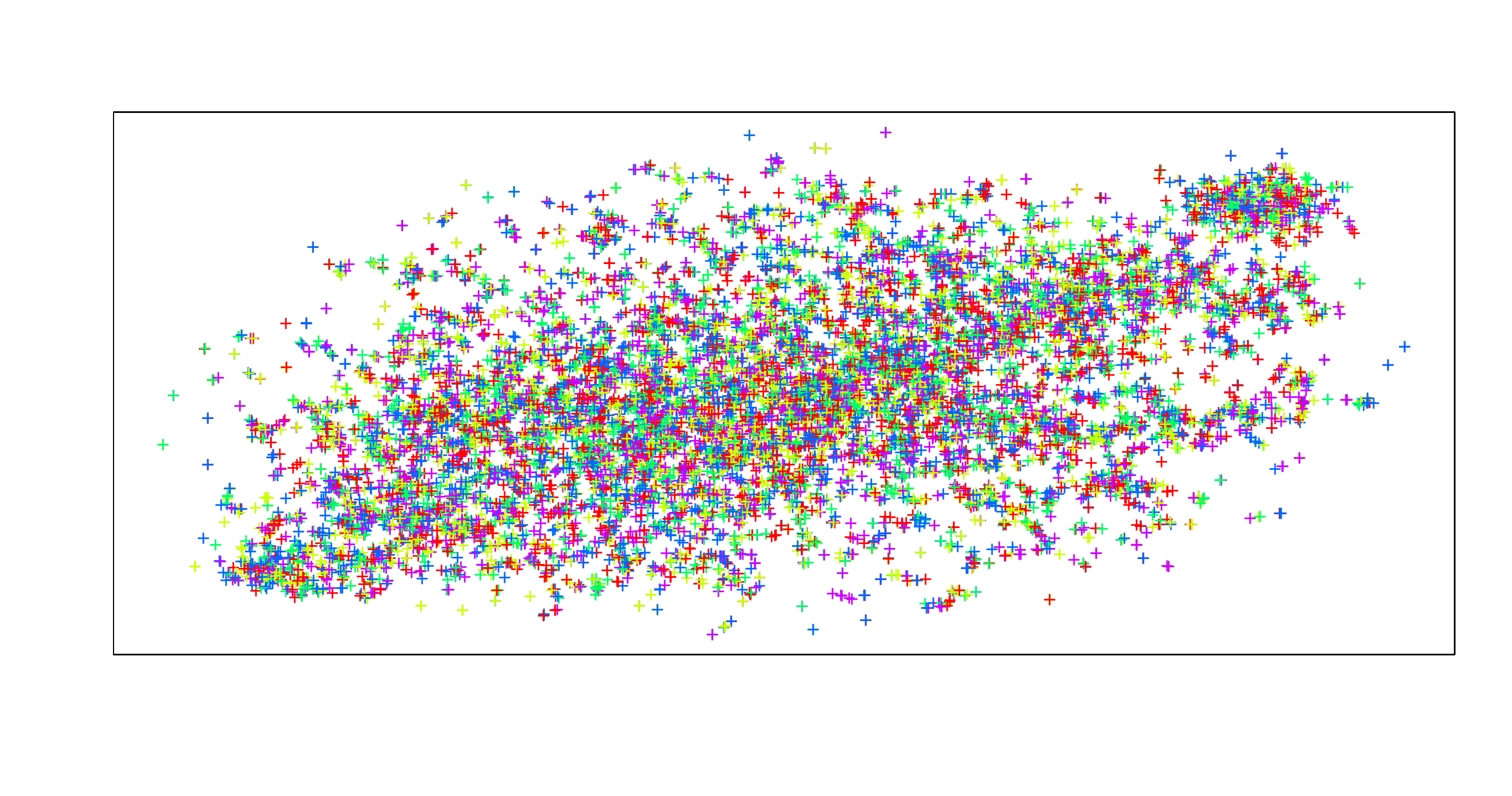}
			\centering\caption[The tSNE for dimensionality reduction of the Epileptic Seizure data]{The tSNE for dimensionality reduction of the Epileptic Seizure data for $1000$ iteration, perplexity = $15$ and theta = $0.5$.}
			\label{fig:3.8}
		\end{minipage}
		\hfil
		\begin{minipage}[b]{0.45\textwidth}
			\includegraphics[width = 3in, height=2in]{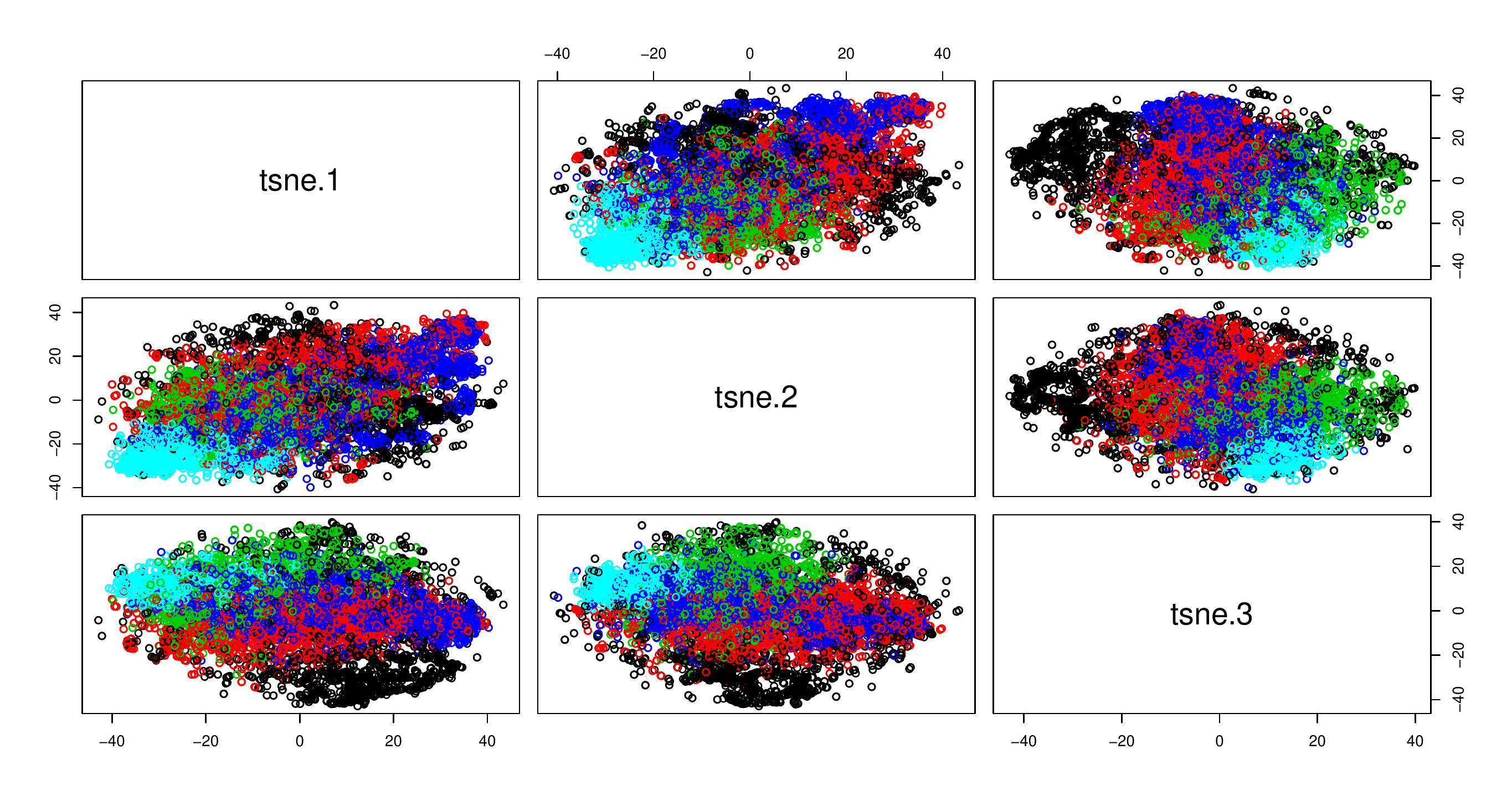}
			\centering\caption[The CWM-tSNE plot for clustering the low-dimensional data produced by tSNE for seizure data with five categories]{The CWM-tSNE plot for clustering the low-dimensional data produced by tSNE for Seizure recognition data with five categories.} 
			\label{fig:3.9}
		\end{minipage}
	\end{figure}
	\vspace{0.05in}
	\bigskip
	\subsection{Epileptic Seizure Recognition}
	We now analyze the Epileptic Seizure recognition data gotten from UCI. The original dataset consists of $5$ different folders, each with $100$ files, with each file representing a single subject/person. Each file is a recording of brain activity for $2.36$ seconds. The corresponding time-series is sampled into $4097$ data points. Each data point is the value of the EEG recording at a different point in time. So there is a total of $500$ individuals with each having $4097$ data points for $23.5$ seconds. Every $4097$ data points is divided and shuffled into $23$ chunks, and each chunk contains $178$ data points for $1$ second. Each data point is the value of the EEG recording at a different point in time. So there is a total of $11500$ pieces of information (row), each with $178$ data points for $1$ second (column), then the last column represents the class $y = \{1,2,3,4,5\}$. The Epileptic data contains $178-$dimensional input vector. The dependent variable $y$ is defined as follows;
	$5$: eyes open when the EEG signal of the brain was recorded. $4$: means eyes closed when the EEG signal was recorded, $3$: mean they identified where the region of the tumor was in the brain and the recorded the EEG activity from the healthy brain area, $2$: means the EEG was recorded from the area where the tumor was located, and $1$: means the recording of seizure activities. The goal is to detect the underlying component of the data. In the previous works, the data has been treated as a binary classification where class $1$ represents the presence of seizure in a patient and $2,3,4,5$ represent the absence of seizure. The label class is distributed equally as $2300$.
	\par The CWMs employs the Ordinary Least squares (OLS) for its maximization step of the EM algorithm, therefore it becomes inappropriate to fit the dependent variable which is a categorical variable. An alternative approach is to take the logarithm of the label class and add some noise to make it a continuous variable. Afterwards, we performed the dimensionality reduction on the independent variable of order $178$. We note here again that the goal of tSNE is not for clustering, however we prioritize dimensionality reduction over clustering with tSNE. Figure (\ref{fig:3.8}) visualizes the high-dimensional data on a $2D$ plane with the $\text{perplexity} = 15$, $\text{iteration} = 1000$ and the $\text{theta} =0.5$. According to the plot shown in Figure (\ref{fig:3.8}), there is a linear pattern as revealed by the tSNE. We observed that when the perplexity is between $9$ and $15$, and the $\text{theta} = 0.5$, tSNE gives an unsatisfactory low-dimensional data, this is called a "crowd point". However, due to high volume of the data, tSNE tends to be a bit slower than when performed on a moderately high-dimensional data. According to the setup of tSNE, there is a trade-off between speed and accuracy. The hidden structure in the high-dimensional data is preserved in the low-dimensional space. However, the epileptic seizure data is highly overlapped, this makes clustering extremely difficult to perform. Figure (\ref{fig:3.9}) shows a five-component structure of the CWMs plot on the low-dimensional data filtered into the CWMs model. Almost all the information criteria selected model with $5$ mixture components. Although, we are able to visualize the high-dimensional data but the clusters are not well separated. One limitation associated with the tSNE output in Figure (\ref{fig:3.8}) is that the information criteria tend to favor the number of the label class. This is however contrary to previous works which have performed binary classification where class $1$ represent presence of Epileptic seizure in the patients against the absence of Epileptic seizure. To reduce the crowd points in Figure (\ref{fig:3.8}), we further performed 
	\begin{figure}[H]
		\centering
		\begin{minipage}[b]{0.45\textwidth}
			\includegraphics[width = 3in, height=2in]{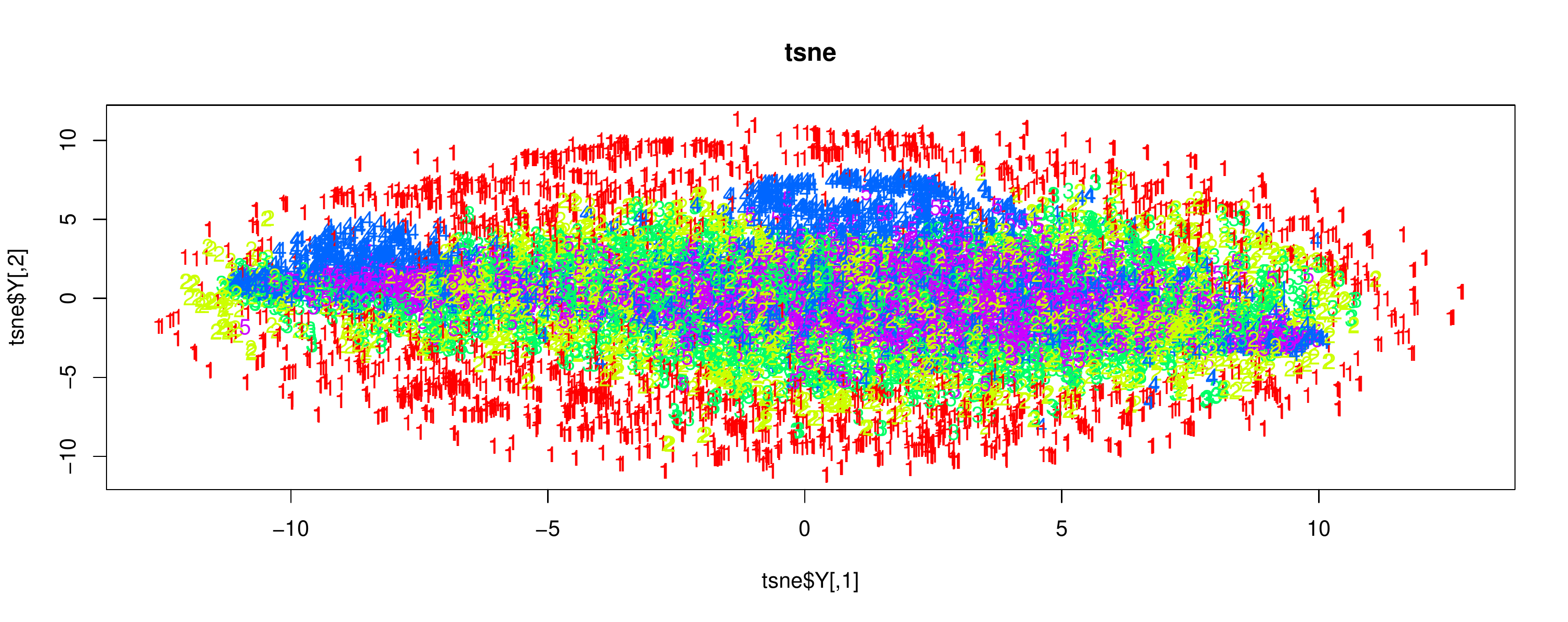}
			\centering\caption[The tSNE for dimensionality reduction of the Epileptic Seizure data]{The tSNE for dimensionality reduction of the Epileptic Seizure data for $10,000$ iterations, perplexity = $250$, and theta = $0.5$.}
			\label{fig:3.10}
		\end{minipage}
		\hfil
		\begin{minipage}[b]{0.45\textwidth}
			\includegraphics[width = 3in, height=2in]{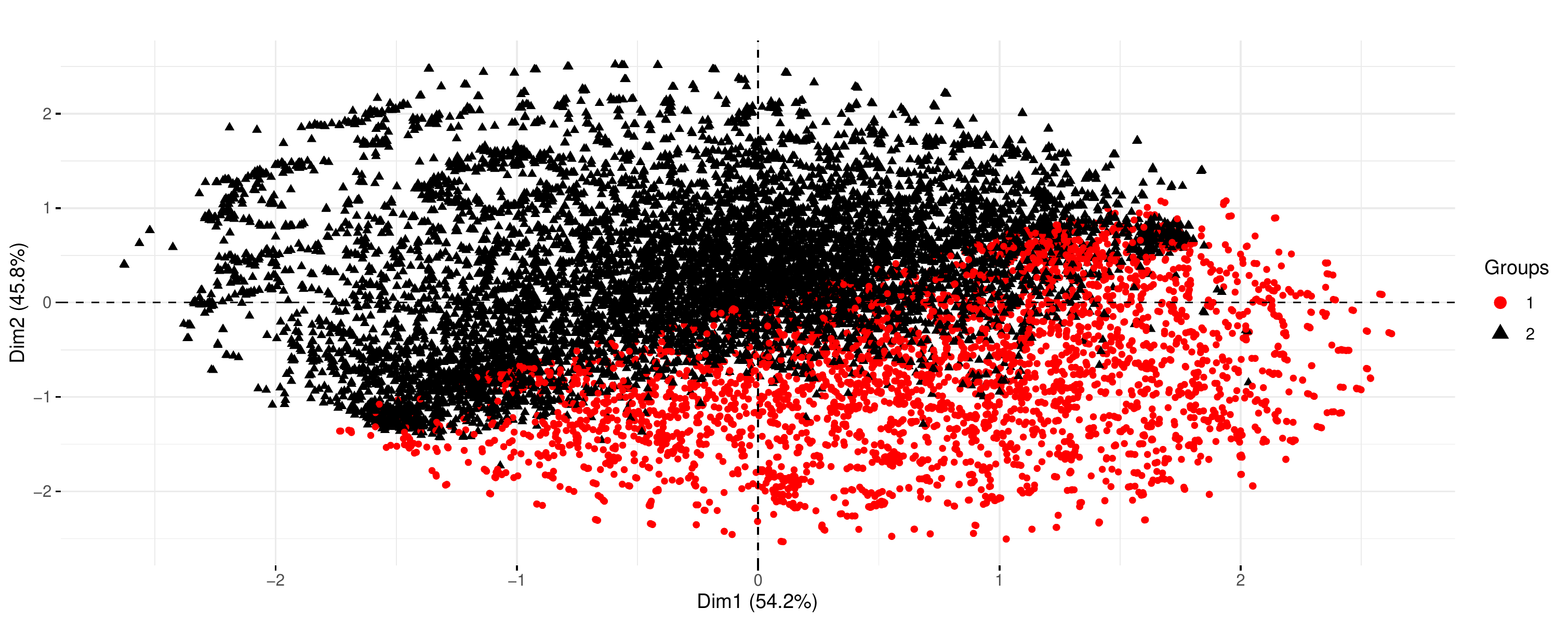}
			\centering\caption[The CWM-tSNE plot for clustering the low-dimensional data produced by tSNE with EEE model with two categories]{The CWM-tSNE plot for clustering the low-dimensional representation of Seizure recognition data produced by tSNE with EEE model.} 
			\label{fig:3.11}
		\end{minipage}
	\end{figure}
	\vspace*{0.05in} 
	\bigskip 
	\noindent a thorough dimensionality reduction with different parameters of the tSNE; the $\text{Perp} = 250$, $\text{theta} = 15$, with $10,000$ iterations. The output after $10,000$ iterations is presented in Figure (\ref{fig:3.10}). From the plot in Figure (\ref{fig:3.10}), the underlying structure was revealed after $10,000$ iterations but tSNE alone is not strong enough to cluster the label class into two classes. CWM-tSNE however worked on the output of the tSNE to reveal the hidden $2$-categorical structure in the seizure data. The plot of CWM-tSNE is shown in Figure (\ref{fig:3.11}). This is the plot produced by the model EEE selected by ICL. CWM-tSNE produces a distinct two classes but with some misclassifications. In Figure (\ref{fig:3.11}), $1$ represents the presence of Epileptic seizure and $2$ represents the absence of the Epileptic seizure. 
	\par
\begin{figure}[H]
		\centering
		\begin{minipage}[b]{0.45\textwidth}
			\includegraphics[width = 3in, height=2in]{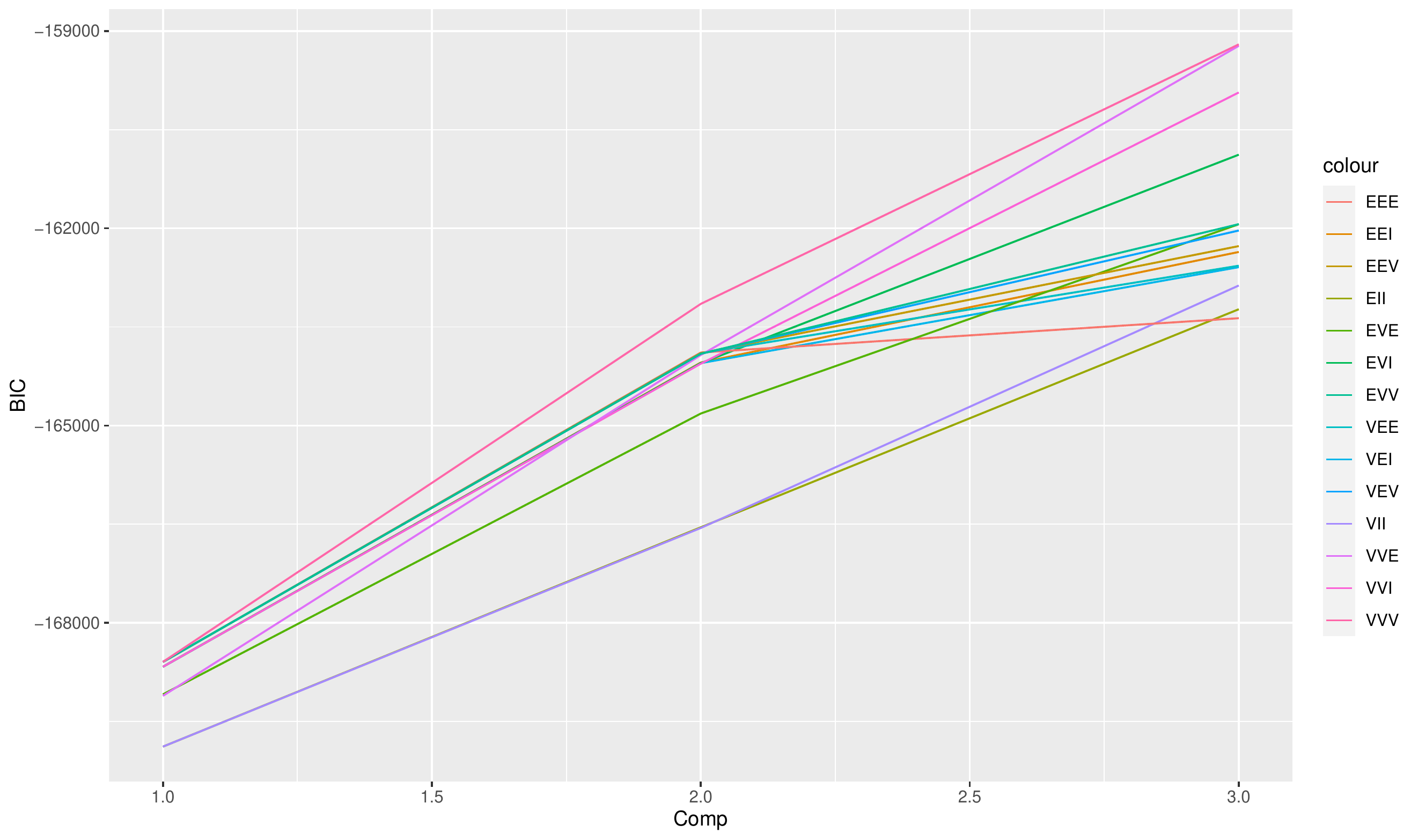}
			\centering\caption[The Model selected by BIC to reveal the hidden component in seizure data]{The Model selection of BIC for Seizure data among the fourteen parsimonious model; BIC selected wrong number of mixture component when the true component according to the label is two-categorical.}
			\label{fig:3.12}
		\end{minipage}
		\hfil
		\begin{minipage}[b]{0.45\textwidth}
			\includegraphics[width = 3in, height=2in]{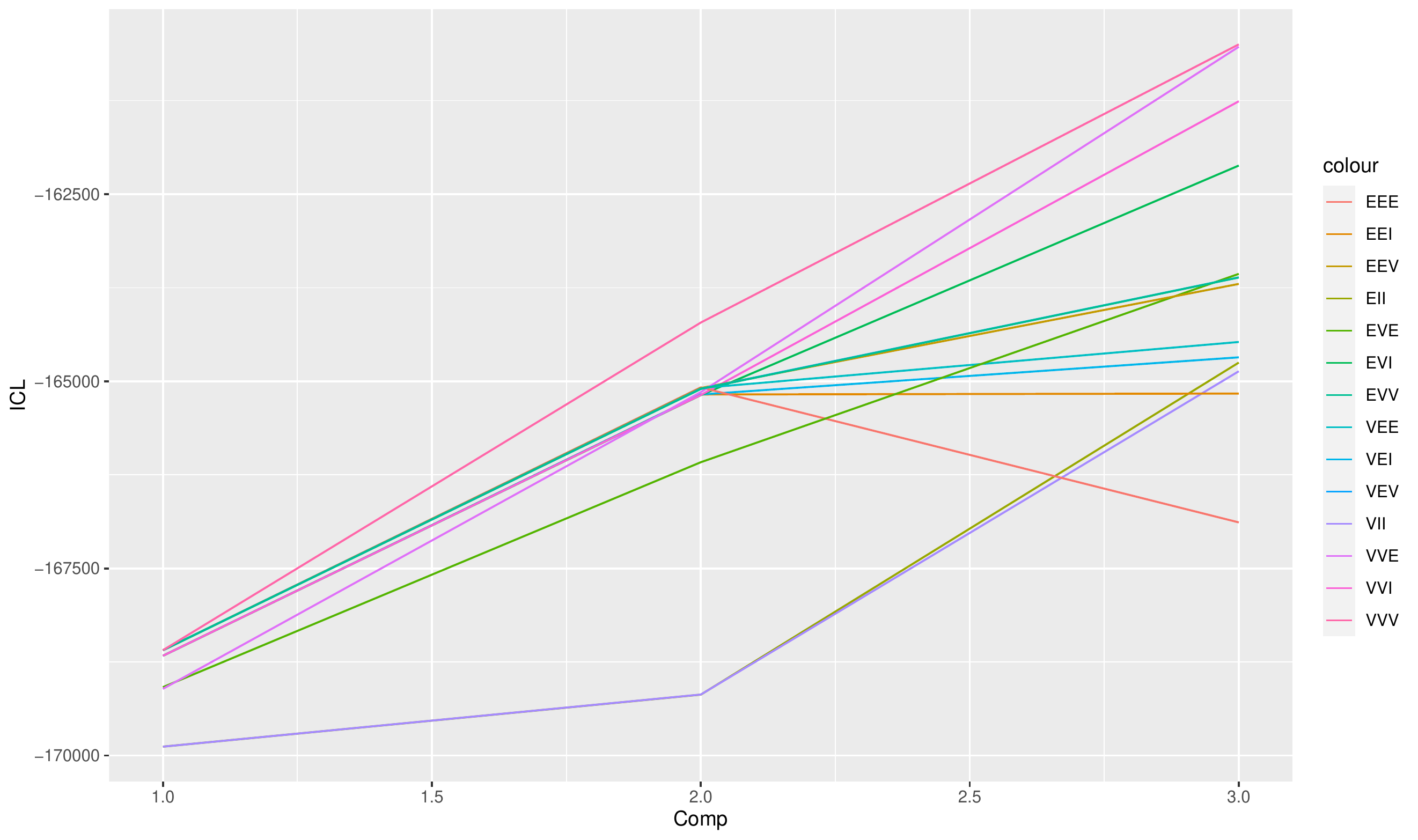}
			\centering\caption[The Model selected by ICL to reveal the hidden component in seizure data]{The Model selection of ICL for Seizure data among the fourteen parsimonious model; ICL selected EEE with the correct number of mixture component when the true component is two categories.} 
			\label{fig:3.13}
		\end{minipage}
	\end{figure}

 \noindent The number of components selected by BIC does not agree with one selected by ICL when using the model EEE. BIC suggested that the number of components is $3$, while ICL suggested that the hidden number of components is $2$. In the other models, the number of components selected by BIC agreed with ICL as they all selected $3$ mixture components. Figure (\ref{fig:3.12}) and Figure (\ref{fig:3.13}) show the comparison between BIC and ICL on the number of mixture components. The values are provided in the Table (\ref{tab:3.7}). The left values are produced by BIC and the right values are the ICL. The ARI and its variants are provided in Table (\ref{tab:3.8}). The model with the highest values of ARI is EVE model. However, the classification accuracy produced by EEE model is $73\%$.
	\vspace*{0.05in} 
	\begin{table}[H]
		\centering\caption[The comparison of the BIC and ICL produced by the fourteen models]{The comparison of the BIC and ICL produced by the fourteen parsimonious models after performing the dimensionality reduction}\vspace{0.4cm}
		\label{tab:3.7}
		\begin{tabular}{r@{\hspace*{.3in}}r@{\hspace*{.3in}}r@{\hspace*{.3in}}r|r@{\hspace*{.3in}}r@{\hspace*{.3in}}r@{\hspace*{.3in}}}
			\hline
			Model&comp1&comp2&comp3&comp1&comp2&comp3\\ 
			\hline
			EII&-169881&-166546&-163231&-169881&-169186&-164750\\
			VII&-169881&-166556&-162870&-169881&-169185&-164866\\
			EEI&-168667&-164041&-162360&-168667&-165176&-165164\\
			VEI&-168667&-164050&-162591&-168667&-165176&-164680\\
			EVI&-168667&-164050&-160881&-168667&-165185&-162118\\
			VVI&-168667&-164059&-159935&-168667&-165183&-161259\\
			EEE&-168593&-163889&-163367&-168593&-$\vec{165081}$&-166885\\
			VEE&-168593&-163898&-162569&-168593&-165090&-164475\\
			EVE&-169086&-164816&-161938&-169086&-166081&-163565\\
			EEV&-168593&-163898&-162271&-168593&-165090&-163699\\		
			VVE&-169109&-163927&-159226&-169109&-165145&-160531\\	
			VEV&-168593&-163908&-162034&-168593&-165099&-163612\\
			EVV&-168593&-163908&-161937&-168593&-165103&-163613\\
			VVV&-168593&-163149&-159204&-168593&-164215&-160499\\
			\hline
		\end{tabular}
	\end{table}
	\begin{table}[H]
		\centering\caption[The comparison of the varieties of ARI produced by the fourteen parsimonious models]{Adjustment Rand Index and its variants of the fourteen parsimonious models to select the hidden structure or cluster in the protein data}\vspace{0.4cm}
		\label{tab:3.8}
		\begin{tabular}{r@{\hspace*{.3in}}r@{\hspace*{.3in}}r@{\hspace*{.3in}}r@{\hspace*{.3in}}r@{\hspace*{.3in}}r@{\hspace*{.3in}}}
			\hline
			Model&Rand&HA&MA&FM&Jaccard\\ 
			\hline
			EII&0.557&0.153&0.153&0.618&0.432\\
			VII&0.477&0.052&0.053&0.520&0.328\\
			EEI&0.503&0.083&0.083&0.555&0.363\\
			VEI&0.474&0.053&0.053&0.516&0.323\\
			EVI&0.708&0.428&0.428&0.759&0.596\\
			VVI&0.486&0.071&0.071&0.529&0.336\\
			EEE&0.601&0.152&0.152&0.686&0.520\\
			VEE&0.478&0.056&0.056&0.522&0.329\\
			EVE&0.712&0.434&0.434&0.763&0.601\\
			EEV&0.539&0.152&0.153&0.589&0.394\\
			VVE&0.486&0.069&0.069&0.529&0.336\\
			VEV&0.509&0.106&0.107&0.556&0.360\\
			EVV&0.705&0.421&0.421&0.756&0.592\\
			VVV&0.485&0.068&0.068&0.529&0.336\\
			\hline
		\end{tabular}
	\end{table}
	\section{Conclusion and Future Work}
	In this paper, we investigated the use of CWMs model on moderately high-dimensional and extremely high-dimensional data. First, we reviewed the general background study of the CWMs according to \cite{Ingrassiaetal2014} and explained how they metamorphosed from a finite mixture model (FMM). 
	According to \cite{Hennig2000}, the problem associated with FMM is the assumption of assignment independence, i.e. the assignment of the data points to the cluster has to be independent of the covariates.
	\par On the contrary, CWMs assume random covariates with a parametric specification which allows for assignment dependence. We further derived the EM algorithm for the parameter estimations. The limitations of CWMs are the main motivation of the paper. The limitation of CWMs is the effect of the "curse of dimensionality". The clustering performance of CWMs is hampered by the dimensionality of the data. However, the eigenvalue decomposition only has a little improvement in the face of huge high-dimensional data. For example, the seizure data of $178$ dimensions has $128,879$ parameters to estimate. This may be impractically attainable in real time when using CWMs, unlike RandomForest that performs internal feature selection. However, the use of eigenvalue decomposition only solves the problems in-part by a little reduction in the number of parameters to be estimated. In the presence of high-dimensional data with CMWs, denegeracies are inevitable, misinterpretation is bound to occur, the computation time increases proportionally with the dimensionality of the data, and low classification performance. For example, an original CWMs fails to cluster an image data with $784$-dimensions.
	\par To alleviate these limitations in CWMs, we introduce a CWMs based on TSNE for high-dimensional data. TSNE is a very powerful dimensionality reduction technique introduced by \cite{maatenandhinton2008}. We first performed a dimensionality reduction based on different parameters of $\mathbf{Rtsne}$ package in $\mathbf{R}$. The approach called CWMs-TSNE is applied to real high-dimensional Epileptic Seizure recognition data. The goal primarily is to detect the hidden mixture component different from the class labels. We investigated different perplexities and selected the one with a satisfactory low-dimensional output. At first, perplexities between $9$ and $15$ gave an unsatisfactory representation with "crowd points" presented in Figure (\ref{fig:3.8}). We further increased the perplexity to $250$. This however contradicts the suggestion given by the authors but the output gave a clear structure. The output however fails to reveal the hidden cluster of the epileptic patients even after $10,000$ iterations [Figure (\ref{fig:3.10})]. Afterwards, the output with the $\text{perplexity} = 250$ was filtered into the CWMs model. At this junction, we applied the $14$ parsimonious models, and we observed a varying computation time due to their varying model complexities. The model selection was performed through eight different information criteria. We observed that the number of mixture component selected BIC did not agree with ICL. While the BIC selected the models with wrong number of components, ICL selected the model EEE with the correct number of hidden components. The output is provided in Figure (\ref{fig:3.11}). However, the overlap reduced drastically when compared to Figure (\ref{fig:3.10}). The data we have used in this Chapter are categorical data with class label more than two classes. All the class labels are first transformed to be continuous variables. This is necessary because the linear Gaussian CWMs models uses OLS for the maximization step and it can only handle a continuous dependent variable efficiently. The possible future direction should be to create a self-sufficient CWMs by embedding a dimensionality reduction technique into the $\mathbf{CWMs}$ package in $\mathbf{R}$. This will allow the package to handle high-dimensional data. In one of my papers, we have tackled the limitation of the family of CWMs and mitigate the effect of the 'curse of dimensionality' on CWMs by developing an appropriate model that is suitable for categorical data in high-dimensional space.
	\bibliographystyle{apalike}
	\bibliography{cwm}
\end{document}